\documentclass{article}

\usepackage{microtype}
\usepackage{graphicx}
\usepackage{subfigure}
\usepackage{booktabs} 

\usepackage{hyperref}

\usepackage[accepted]{icml2024}

\usepackage{amsmath}
\usepackage{amssymb}
\usepackage{mathtools}
\usepackage{amsthm}

\usepackage[capitalize]{cleveref}

\theoremstyle{plain}
\newtheorem{theorem}{Theorem}[section]

\theoremstyle{definition}
\newtheorem{definition}[theorem]{Definition}

\theoremstyle{remark}

\usepackage[textsize=tiny, disable]{todonotes}

\usepackage{lipsum}

\usepackage{times}
\usepackage[utf8]{inputenc} 
\usepackage[T1]{fontenc}    
\usepackage{url}            
\usepackage{booktabs}       
\usepackage{nicefrac}       
\usepackage{microtype}      
\usepackage{graphics,color}

\usepackage{algorithm}
\usepackage{algorithmic}
\usepackage{enumitem}

\usepackage{amsfonts}       
\usepackage{amsmath}       
\usepackage{amssymb}

\newcommand{\figref}[1]{Fig.~\ref{#1}}    

\newcommand{\tabref}[1]{Table~\ref{#1}}

\newcommand{\fs}[1]{{\bf #1}}

\usepackage{xspace}
\makeatletter
\DeclareRobustCommand\onedot{\futurelet\@let@token\@onedot}
\def\@onedot{\ifx\@let@token.\else.\null\fi\xspace}
\makeatother

\newcommand{\ie}{i.e\onedot}

\usepackage{xr-hyper}
\makeatletter
\newcommand*{\addFileDependency}[1]{
  \typeout{(#1)}
  \@addtofilelist{#1}
  \IfFileExists{#1}{}{\typeout{No file #1.}}
}
\makeatother

\usepackage{xcolor}
\definecolor{ourblue}{rgb}{0.368,0.507,0.71}
\definecolor{ourorange}{rgb}{0.881,0.611,0.142}
\definecolor{ourgreen}{rgb}{0.56,0.692,0.195}
\definecolor{ourred}{rgb}{0.923,0.386,0.209}
\definecolor{ourviolet}{rgb}{0.528,0.471,0.701}
\definecolor{ourbrown}{rgb}{0.772,0.432,0.102}
\definecolor{ourlightblue}{rgb}{0.364,0.619,0.782}
\definecolor{ourdarkolive}{rgb}{0.572,0.586,0.}
\definecolor{ourdarkred}{rgb}{0.67, 0.22, 0.07}
\definecolor{ourdarkorange}{rgb}{0.71, 0.49, 0.1}
\definecolor{ourdarkblue}{rgb}{0.27, 0.4, 0.58}
\definecolor{ourdarkgreen}{rgb}{0.41, 0.51, 0.15}

\definecolor{ourcyan2}{rgb}{0.125,0.722,0.804}
\definecolor{ourred2}{rgb}{0.863,0.184,0.047}
\definecolor{ouryellow2}{cmyk}{0,0.16,1.0,0.07}
\definecolor{ourviolet2}{cmyk}{0.55,0.56,0,0.47}
\definecolor{ourorange2}{cmyk}{0,0.46,0.89,0.11}

\usepackage{multirow}

\newcommand{\indep}{\perp \!\!\! \perp}
\newcommand{\algo}[1]{\textsc{#1}}
\newcommand{\method}{\algo{CAIAC}\xspace}

\def\code#1{\texttt{#1}}
\newcommand{\STAB}[1]{\begin{tabular}{@{}c@{}}#1\end{tabular}}

\newcounter{inlineequation}
\renewcommand{\theinlineequation}{(\Roman{inlineequation})}

\newcommand{\inlineeq}[1]{\refstepcounter{inlineequation}\theinlineequation\ \(#1\)}

\usepackage{amsmath,amsfonts,bm}

\def\1{\bm{1}}

\DeclareMathAlphabet{\mathsfit}{\encodingdefault}{\sfdefault}{m}{sl}
\SetMathAlphabet{\mathsfit}{bold}{\encodingdefault}{\sfdefault}{bx}{n}

\def\gA{{\mathcal{A}}}

\def\gD{{\mathcal{D}}}

\def\gG{{\mathcal{G}}}

\def\gP{{\mathcal{P}}}

\def\gR{{\mathcal{R}}}
\def\gS{{\mathcal{S}}}

\def\gU{{\mathcal{U}}}
\def\gV{{\mathcal{V}}}

\def\sR{{\mathbb{R}}}

\newcommand{\E}{\mathbb{E}}

\newcommand{\parents}{\mathrm{Pa}} 

\usepackage{amsmath}
\usepackage{wrapfig}
\usepackage{caption}
\usepackage{subcaption}
\usepackage{graphicx}
\usepackage{hyperref}
\usepackage{url}
\hypersetup{
	colorlinks=true,       
	linkcolor=ourdarkblue,        
	citecolor=ourdarkgreen,        
	filecolor=ourdarkblue,     
	urlcolor=ourviolet
}

\newif\ifcomments
\commentstrue 

\newcommand{\doo}{\textit{do}}

\icmltitlerunning{Causal Action Influence Aware Counterfactual Data Augmentation}

\begin{document}

\twocolumn[
\icmltitle{Causal Action Influence Aware Counterfactual Data Augmentation}


\icmlsetsymbol{equal}{*}

\begin{icmlauthorlist}
\icmlauthor{Núria Armengol Urpí}{eth,mpi}
\icmlauthor{Marco Bagatella}{eth,mpi}
\icmlauthor{Marin Vlastelica}{mpi}
\icmlauthor{Georg Martius}{unitubi,mpi}
\end{icmlauthorlist}

\icmlaffiliation{eth}{Department of Computer Science, ETH Zurich, Zurich, Switzerland}
\icmlaffiliation{mpi}{Max Planck Institute for Intelligent Systems, Tübingen, Germany}
\icmlaffiliation{unitubi}{Department of Computer Science, University of Tübingen, Tübingen, Germany}

\icmlcorrespondingauthor{Núria Armengol Urpí}{nuriaa@ethz.ch}

\icmlkeywords{causal action influence, offline, deep reinforcement learning, data augmentation, learning from demonstrations, out-of-distribution generalization}

\vskip 0.3in
]



\printAffiliationsAndNotice{}  

\begin{abstract}

\looseness -1 Offline data are both valuable and practical resources for teaching robots complex behaviors.
Ideally, learning agents should not be constrained by the scarcity of available demonstrations, but rather generalize beyond the training distribution.
However, the complexity of real-world scenarios typically requires huge amounts of data to prevent neural network policies from picking up on spurious correlations and learning non-causal relationships. 
We propose \method, a data augmentation method that can create feasible synthetic transitions from a fixed dataset without having access to online environment interactions.
By utilizing principled methods for quantifying causal influence, we are able to perform counterfactual reasoning by swapping \emph{action}-unaffected parts of the state-space between independent trajectories in the dataset.
We empirically show that this leads to a substantial increase in robustness of offline learning algorithms against distributional shift.
Videos, code and data are available at \url{https://sites.google.com/view/caiac}.

\end{abstract}

\section{Introduction}


Offline learning offers the opportunity of leveraging plentiful amounts of prerecorded data in situations where environment interaction is costly \citep{bahl2022human, brohan2022rt, brohan2023rt,vlastelica2023diverse}.
However, one of the fundamental challenges of such a framework is that of causal confusion.

\begin{figure}
    \includegraphics[width=\linewidth]{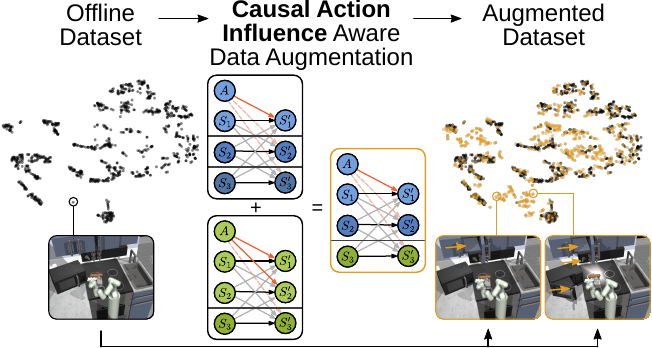}
    \vspace{-0mm}
    \captionof{figure}{Overview of the proposed approach. 
    Interactions between the agent and entities in the world are sparse. 
    We use causal action influence (CAI), a local causal measure, to determine action-independent entities and create counterfactual data by swapping states of these entities from other observations in the dataset. Offline learning with these augmentations leads to better generalization.
    } \label{fig:overview}
    \vspace{-4mm}
\end{figure}

Causal confusion arises when a trained agent misinterprets the underlying causal mechanics of the environment and, hence, fails to distinguish spurious correlations from causal links~\citep{de2019causal, gupta2023can}. 
When trying to reduce training loss, an agent can benefit from such spurious correlations in the data and, therefore, they can be inadvertently transferred to the mechanisms of learned models ~\citep{gupta2023can}. 

Problematically, causally confused agents, are prone to catastrophic failure even in mild cases of distributional shift~\citep{de2019causal}, \ie when the test distribution deviates from the training distribution.
Subtle forms of distributional shift are common when learning from real-world data: collected demonstrations can only encompass a small fraction of the vast amount of possible configurations stemming from the inherent presence of many entities in the real world \citep{battaglia2018relational}.
Hence, at test times, agents are often queried on unfamiliar (\ie out of distribution) configurations.



To illustrate the problem, let us imagine that we have demonstrations of a robot performing several kitchen-related activities: opening a microwave, sliding cabinets and turning on and off a light switch.
However, the demonstrations only show how to perform those activities in this pre-specified order. Hence, following the example, the sliding cabinet (Event Y) is only shown to be open when the microwave (Event X) is open as well.\looseness-1

\begin{figure}
    \begin{minipage}{0.5\linewidth}
        \centering
        \includegraphics[width=\linewidth]{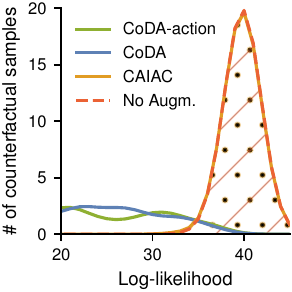}
    \end{minipage}%
    \begin{minipage}{0.5\linewidth}
    \centering
        \includegraphics[width=\linewidth]{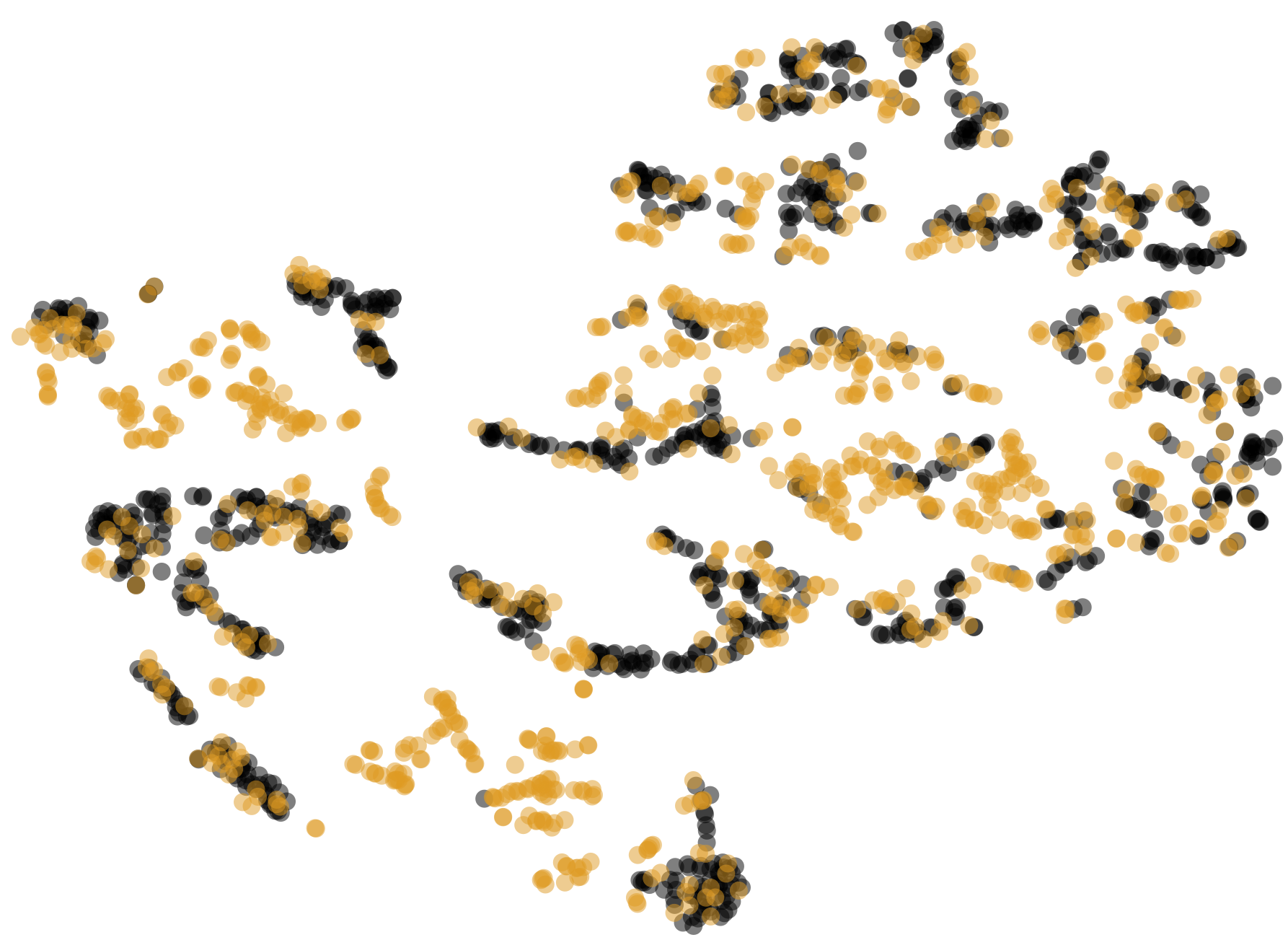}
    \end{minipage}
        \vspace{-0mm}
    \caption{ \method counterfactual samples are consistent with the environment's dynamics and increase the support of the joint state space distribution, enabling the agent to be robust to distributional shift. Left: Log-likelihoods under the environment transition kernel of counterfactuals created with different methods. Right: Original data and counterfactuals augmentations with \method visualized with t-SNE. Details on this evaluation are reported in Appendix \ref{app:test_quality_cf}.}
    \vspace{-0mm}
    \label{fig:quality_cf}
\end{figure}

In this scenario, the trained agent happens to \textit{simultaneously} observe independent events X and Y whenever it executes the sliding action (A), and hence may attribute the action A to X and Y occurring jointly, even though event X is independent of A.
If the spurious correlation between X and Y observed during training  fails to persist at test time, such a causally confused model may exhibit subpar performance. Namely, the agent might not be able to slide open the cabinet when the microwave is closed.

In contrast, humans are remarkably good at inferring what parts of the environment are relevant to solve a task, possibly due to relying on a causal representation of the world \citep{pearl2018book}.
This hypothesis has motivated the creation of \textit{causal} approaches in machine learning that aim to identify relationships in the environment that will remain invariant under changes in the data-generating distribution.
Existing work at the intersection between RL and causality has focused on an online learning \citep{lyle2021resolving, wang2022causal, ding2023seeing}, imitation learning \citep{de2019causal} or partial observability setting \citep{forney2017counterfactual, kallus2018confounding}.
When learning in the online setting, some works operate in the interventional setting \citep{lyle2021resolving, de2019causal}, \ie a user may be able to "experiment" in the environment in order to discover causal structures by assigning values to some subset of the variables of interest and observing the effects on the rest of the system.
In contrast, we focus on the challenging \textit{offline} setting, where the agent is \textit{not} capable of observing the real effects of such an intervention, and we propose an observational approach.
While one could also learn to predict the outcome of the interventions by using a model-based approach, we note that an uninformed dynamics model can still  be sensitive to spurious correlations and suffer from approximating errors.


Our approach,  which we refer to as \emph{\textbf{Ca}usal \textbf{I}nfluence \textbf{A}ware \textbf{C}ounterfactual Data Augmentation} (\method), introduces counterfactual data augmentations without the need for additional environment interactions, or reliance on counterfactual model rollouts.
Instead, we exploit collected data to learn a causal model that explicitly reasons about causal influence and swap locally causally independent factors across different observed trajectories. Estimating the entire causal structure remains, however, a challenging task, particularly if attempted from offline data. 

\looseness -1 Taking this into account, we focus on identifying the effects the agent has on the environment; we thus assume action-influence to be more important for policy learning than potential object-object interactions.
By partially trading off generality, this inductive bias on the underlying causal structure reduces the problem of estimating the full causal structure to only measuring the influence of actions over objects.
Although causal discovery from observational data is known to be impossible for the general case~\citep{pearl2009causality, PetersJanzingSchoelkopf17}, methods exist that make use of some form of independence testing that have been successful in applications.



While other causal methods rely on heuristics to create new samples \citep{ding2023seeing}, on implicit measures for detecting causal influence among entities \citep{Pitis2020} or on regularization of the dynamical models to suppress spurious correlations \citep{ding2023seeing,wang2021task}, our method is theoretically sound and relies on an \textit{explicit} measure of influence, namely state-conditioned mutual information \citep{cover1999elements}.
We find this approach to be significantly more reliable at creating counterfactual samples that, not only follow the environment's dynamics (i.e. they are feasible), but also increase the support of the joint distribution over environment entities, as shown in \figref{fig:quality_cf}.


Moreover, we show that by providing an offline learning agent with \method 's counterfactual samples, we prevent the agent from suffering from causal confusion, and we hence improve robustness to distributional shifts at test time.
Our framework works as an independent module and can be used with any data-driven control algorithm.
We demonstrate this through empirical results in high-dimensional offline goal-conditioned tasks, applying our method to fundamentally different data distributions and learning methods.
Namely, we couple our method with offline goal-conditioned skill learning on the Franka-Kitchen environment \citep{gupta2019}, and classical offline goal-conditioned reinforcement learning on Fetch-Push and FetchPick\&Lift \citep{plappert2018multi}. 
Across experiments, we show that \method  leads to enhanced performance in out-of-distribution settings and when learning from a modest amount of demonstrations. 

\section{Background}

\subsection{Reinforcement Learning}
\looseness -1 Markov Decision Processes (MDPs) are used as the basic formalism for sequential decision-making problems.
\begin{definition}[Markov Decision Process (MDP)]
A Markov Decision Process is a tuple $(\mathcal{S,A}, \gP, \gR, \rho_0, \gamma)$, consisting of state space $\gS$, action space $\gA$, transition kernel $\gP(S' \mid\, S, A)$, reward function $\gR: \gS \times \gA \mapsto \sR$, initial state distribution $\rho_0$ and discount factor $\gamma$, respectively.    
\end{definition}
The goal of a learning algorithm is to extract a policy $\pi: \gS \mapsto \gA$ for maximizing the expected return $\E_\pi[\sum_{t=0}^\infty \gamma^t r(s_t, a_t)]$.
Online algorithms have access to samples from the transition kernel $\gP$, while offline methods leverage a fixed dataset $\gD$ of trajectories which may be suboptimal. 
In this work, we focus on the offline setting, which particularly suffers from issues of distributional shift and out-of-distribution generalization.

\subsection{Causal Graphical Models}
\label{subsec:causal_graphical_models}
Generally, a joint distribution $P(X)$ has a particular independence structure which induces a specific factorization.
This independence structure is a consequence of the functional relationship (also called mechanism) between the variables that can be accurately described through a Structural Causal Model (SCM).
\begin{definition}[SCM \citep{pearl2009causality}] A SCM is a tuple $(\mathcal{U}, \mathcal{V}, F, P^u)$, where $\mathcal{U}$ is a set of exogenous (unobserved) variables (e.g. the unobserved source of stochasticity in the environment) sampled from $P^U$, $\mathcal{V}$ is a set of endogenous (observed) variables (e.g. the observed state, the action and the reward in RL). $F$ is the set of structural functions capturing the causal relations, such that functions $f_V: \parents(V) \times \mathcal{U} \rightarrow V$ with $\parents(V) \subset \mathcal{V}$ denoting the set of parents of $V$, determine the value of endogenous variables $V$ for each $V \in \mathcal{V}$.
\end{definition}

SCMs are usually visualized as a directed acyclic graph $\gG$ whose nodes are associated with the variables in the SCM and whose edges indicate causal relationships. 
We say that a pair of variables $v_i$ and $v_j$ are confounded by a variable $C$ (confounder) if they are both caused by $C$, i.e., $C \in \parents(v_i)$ and $C \in \parents(v_j)$. When two variables $v_i$ and $v_j$ do not have a direct causal link, they are still correlated if they are confounded, in which case this correlation is a \textit{spurious correlation}. 
Given an SCM, one can make inferences about causal effects through the concept of an intervention.
\begin{definition}[$\doo$-intervention \citep{pearl2009causality}]
An intervention $\textit{do}(V = v)$ on $V$ induces a new SCM = $(\mathcal{U}, \mathcal{V}, F', P^u)$, where $F' = \{f_W \in F \mid W \neq V  \} \cup \{ f_{V=v} \}$ and $f_{V=v}(p,u)=v \; \forall p \in \parents(V), u \in \mathcal{U}.$
\end{definition}
An intervention on a set of nodes of the SCM effectively changes their structural equations, mostly replacing them by an explicit value.
Interventional queries of the form $P( Y \vert \ do(X=x) )$ are the so-called second rung of causation~\citep{pearl2009causality}. 
In this work we are interested in the third, counterfactual queries. 
\begin{definition}[Counterfactual]\label{def:counterfactual} A counterfactual query is a query of the form  $P(Y \mid\, \doo(X=x), \gU = u)$, where $Y,X \subset \gV$ and $\gU$ is the set of exogenous variables of the underlying SCM.
\end{definition}
A counterfactual query about variables $Y$ is asking what would have happened to $Y$ if under the same conditions $\gU = u$ the intervention $\doo(X=x)$ had been performed.

\section{Problem Definition}
We assume a known and fixed state-space factorization $\mathcal{S}=\mathcal{S}_1  \times ... \times \mathcal{S}_N$ for $N$ entities, where each factor $\mathcal{S}_i$ corresponds to the state of an entity. 
In practice, there are methods that allow to automatically determine the number of factors \citep{Zaheer2017deep} and to learn latent representations of each entity \citep{burgess2019monet,zadaianchuk2023objectcentric,locatello2020object, greff2019multi, jiang2019scalor, seitzer2022bridging}. 
While we do not consider them for simplicity, our method can be applied on top of such techniques.

An MDP coupled with a policy $\pi: \gS \mapsto \gA$ induces an SCM describing the resulting trajectory distribution. 
Given the Markovian property of the MDP and flow of time, there only exist direct causal links $\{S_t, A_t\} \rightarrow S_{t+1}$ by definition, \ie $S_{t+1} \indep V \mid \{S_t, A_t\}$ for non-descendant nodes $V \notin \{S_t, A_t\}$. 
For our purposes, it suffices to look at the time slice sub-graph which is governed by the MDP transition kernel $\gP$ between state $S$, action $A$ and next state $S'$ with state factors $\{S_i\}_{i=1}^N$.

In most non-trivial environments, there exists an edge $S_i / A  \rightarrow S'_j$ for most $i,j$ (\figref{fig:method}(a)).
However, interactions often become sparse once we observe a particular state configuration.
We capture these local interactions by the notion of a \emph{local causal model}.
\begin{definition}[Local Causal Model \citep{Pitis2020}]\label{def:lcm}
Given an SCM $(\mathcal{U}, \mathcal{V}, F, P^u)$, the local SCM induced by observing $V=v$ with $V \subset \mathcal{V}$ is  the SCM with $F_{V=v}$ and the graph $\gG_{do(V=v)}$ resulting from removing edges from $\gG_{do(V=v)}$ until the graph is causally minimal.
\end{definition}
We shall use shorthand $\gG_v$ for $\gG_{\doo(V=v)}$ and similar to reduce notational clutter.
Where normally the vertex set $\{A\} \cup \{S'_j\}_{j=1}^N$ would be densely connected in the direction $A \rightarrow S'$, intervening on $S$ results in a sparser causal dependency in $\gG_s$.
An example of this local causal structure is given in \figref{fig:method}(b): the robot can only influence the kettle and its own end-effector through its actions, but none of the other entities.
We will make heavy use of sparsity and locality in constructing a counterfactual data augmentation.

\section{Method} 

%
The main challenge that we aim to tackle in this work is that of learning policies in the offline regime that are more robust to distributional shift, \ie are not susceptible to spurious correlations.
We achieve this by augmenting real data with counterfactual modifications to \emph{causally action-unaffected} entities, hence creating samples outside the support of the data distribution.
Our method relies on the observation that performing the intervention $do(S = s)$ reduces the number of edges between $A$ and $S'$ as per \cref{def:lcm}, leaving some factors independent of $A$.
Now, assuming that we observe the transition $S=s, A=a, S'=s'$ we pose the question of how we can do counterfactual reasoning without access to the true set of structural equations $F_{S=s}$, \ie synthesize counterfactual transitions.

\begin{figure*}
    \centering
    \vspace{-0mm}
    \includegraphics[width=.98\linewidth]{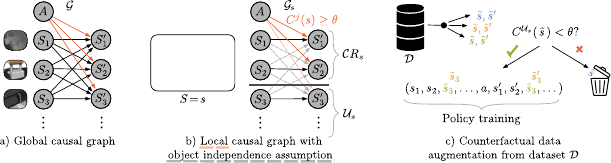}\vspace{-.0em}
    \caption{Illustration of counterfactual data augmentation. 
    The global causal graph does not allow for factorization (a). Our local causal graph (b) is pruned by causal action influence. Object-object interactions are assumed to be rare/not existing (gray dashed).
    We swap elements that are not under control (\ie in set  $\mathcal U$) by samples from the data, thus creating counterfactual samples.
    We omit the exogenous variables from the global graph for compactness.\looseness-1}
    \label{fig:method}
    \vspace{-0mm}
\end{figure*}

To this end, we need to learn the causal structure of the LCM, which is known to be a hard problem  \citep{PetersJanzingSchoelkopf17}.
Therefore, we make the key assumption that interactions between entities are sparse (\ie only occur rarely) and are thus negligible.
While the correctness of generated counterfactuals will rely on this assumption to hold, we argue that this is realistic in various robotics tasks.
For example, in the kitchen environment depicted in \figref{fig:overview}, the entities can hardly influence each other.
In fact, each entity is mostly controlled by the agent actions.
This would also be the case in several manufacturing processes, in which interaction between entities should only happen under direct control of robots.
Moreover, settings in which the assumption does not hold remain a significant challenge for most heuristic methods for causal discovery~\cite{Pitis2020}, which underperform despite their generality.
More formally, and in a graphical sense, we assume that there is no arrow $S_i \rightarrow S'_j, i\neq j$ as visualized by the gray dashed lines in \figref{fig:method}(b).
We note that only two groups of arrows remain in the causal graph: $S_j \rightarrow S'_j$, which we assume to always be present, and $A \rightarrow S'_j$.
This practical assumption allows us to reduce the hard problem of local causal discovery to the more approachable problem of local action influence detection. 
As a result, instead of resorting to heuristics~\citep{Pitis2020}, we make use of a more principled \emph{explicit} measure of local influence, the  Causal Action Influence (CAI)~\citep{Seitzer2021CID} measure, which we introduce below.\looseness-1

\subsection{Causal Action Influence Detection} \label{sec:cai}

To predict the existence of the edge $A \rightarrow S'_j$ in the local causal graph $\gG_s$, \citet{Seitzer2021CID} use conditional mutual information (CMI) \citep{cover1999elements} as a measure of dependence, which is zero if $S'_j \indep A | S=s$. 
Therefore, in each state $S=s$ we use the point-wise CMI as a state-dependent quantity that measures causal action influence (CAI), given by
\begin{align}
\begin{split}
    C^j(s) &:= I(S'_j; A \mid S=s)  \\ &= \mathbb E_{a\sim \pi} \big[ D_{KL} \big( P_{S'_j\mid S=s,A=a}   \;\big|\big| \; P_{S'_j\mid S = s} \big) \big].
    \label{eq:pointwise_cmi}
\end{split}
\end{align}
The transition model $P_{S'_j\mid S=s,A=a}$ is 
modeled as a Gaussian neural network (predicting mean and variance) which maximizes a log-likelihood objective. 
The conditional distribution $P_{S^j_{t+1}\mid S=s}$ is computed in practice using $M$ empirical action samples with the full model: $P_{S'_j|S=s} \approx \frac{1}{M}\sum_{m=1}^{M} P_{S'_j\mid S=s,A=a^{(m)}},\,\, a^{(m)} \sim \pi$. 
The KL divergence in \eqref{eq:pointwise_cmi} can be estimated using an approximation for Gaussian mixtures from \citep{Durrieu2012}.
We note that the transition model does not need to be queried autoregressively, which avoids the issue of compounding errors.
We refer the reader to \citet{Seitzer2021CID} for more details.

\subsection{Inferring Local Factorization}

Having introduced the concepts of locality and object independence, as well as a method to detect causal action influence, we proceed to infer the local factorization which will be leveraged to create counterfactual experience.
For each state $s$ in our data set $\mathcal{D}$, 
 we compute the uncontrollable set, as the set of entities in $s$ for which the agent has no causal action influence, expressed as:
\begin{equation}
    \mathcal{U}_{s} = \{s_j \mid  C^j(s) \leq \theta, j \in [1, N] \} 
    \label{eq:uncontr_set}
\end{equation}
where $\theta$ is a fixed threshold. 
The set $\mathcal{U}_{s}$ contains all entities $j$ for which the arrow $A\rightarrow S'_j$ in the local causal graph $\gG_{s}$ does not exist.
The remaining entities are contained in the set of controllable entities $\mathcal{CR}_{s} = \{s_1,\dots,s_N\} \setminus \mathcal U_s$.
An illustration is given in \figref{fig:method}(b).

With our assumptions and the sets $\mathcal{U}_s$ and $\mathcal{CR}_s$ we find that the local causal graph $\gG_s$ 
is divided into the \emph{disconnected} subgraphs $\gG_s^{\mathcal{CR}}$, that contains the entities in $\mathcal{CR}$ and $A$, and into $|\mathcal{U}_s|$ \emph{disconnected} subgraphs $\gG_{s_i}^{\mathcal U}, i \in [1, |\mathcal{U}_s|]$, each of which contains an entity in $\mathcal{U}_s$ with only self-links, see \figref{fig:method}(b). We can also compute the uncontrollable set for an extended time period $\kappa$, \ie \inlineeq{\mathcal{U}_{s_{t:t+\kappa}} = \bigcap_{\tau=t}^{t+\kappa-1}\mathcal{U}_{s_\tau} }\label{eqn:uncontrollable_intersection}.


\begin{figure}
 \hrule height 1pt \vspace{.1em}   
 \captionof{algorithm}{\method }\vspace{-1em} \label{alg:algorithm}
 \hrule
 {\small
\begin{algorithmic}
    \INPUT Dataset $\mathcal{D}$
    \STATE Compute uncontrollable set $\mathcal{U}_{s},\ \forall s\in \mathcal{D}$ (Eq.~\ref{eq:uncontr_set}).
    \STATE Sample $(s, a, s') \sim \mathcal{D}$ and set  $(\tilde{s}, \tilde s') \leftarrow (s, s')$  
    \STATE \textbf{for} $s_i \in \mathcal{U}_{s}$ 
    \STATE \quad Sample $(\hat s, \hat a, \hat s') \sim \mathcal{D}$
    \STATE \quad \textbf{if} $\hat{s}_i \in \mathcal{U}_{\hat s} $ \textbf{then} $(\tilde{s}_i, \tilde s_i') \leftarrow (\hat{s}_i, \hat s_i')$ 
    \STATE \quad \textbf{yield}  $(s, a, s')$ and $(\tilde s, a, \tilde s')$
\end{algorithmic}
}
 \hrule height 1pt
\vspace*{-0em} 
\end{figure}

\subsection{Computing Counterfactuals}
Given the partitioning of the graph described above, similarly to \citep{Pitis2020}, we can think of each subgraph as an independent causal mechanism that can be reasoned about separately.
Assuming no unobserved exogenous variables, we may obtain counterfactuals in the following way:
given two transitions $(s,a,s')$ and $(\hat s, \hat a, \hat s') \in \mathcal D$ sampled for training, which have at least one uncontrollable subgraph structure in common (\ie $\mathcal{U}_s \cap\, \mathcal{U}_{\hat s} \neq \emptyset$), we generate a counterfactual transition $(\tilde{s}, \tilde{a}, \tilde{s}')$ by swapping the entity transitions $(s_i, s'_i)$  with $(\hat s_i, \hat s'_i)$ and $i \in \mathcal{U}_s \cap\, \mathcal{U}_{\hat s}$.
In this way, we observe the result of the counterfactual query $P(S' \mid \doo(S=\tilde s, A=a))$ without using the mechanism $f_{S'}$ if it remains unchanged in the new LCM.
However, even when only swapping the entity transitions for $\gU_s \cap\, \gU_{\hat s}$, the LCM resulting from the intervention $\doo(S = \tilde{s})$ may still contain a different mechanism $f_{S'}$ than the source LCM of $s$, meaning that the transition $(\tilde{s}, \tilde{a}, \tilde{s}')$ becomes invalid.
An additional check of causal influence would entice an out-of-distribution query to the CAI measure, which is learned from transitions and therefore error-prone.
In practice we do not perform this check and accept creating a small fraction of potentially infeasible transitions. 
The pseudocode of our method, which we call \textbf{Ca}usal \textbf{I}nfluence \textbf{A}ware \textbf{C}ounterfactual Data Augmentation (\method), is given in Algorithm \ref{alg:algorithm}.  



\section{Related work}

\fs{Data Augmentation}
Data augmentation is a fundamental technique for achieving improved sample-efficiency and generalization to new environments, especially in high-dimensional settings.
In deep learning systems designed for computer vision, data augmentation can be found as early as in \citet{lecun1998gradient} and \citet{krizhevsky2012imagenet}, who leverage simple geometric transformations, such as random flips and crops.
Naturally, a plethora of augmentation techniques \citep{berthelot2019mixmatch, sohn2020fixmatch} have been proposed over time.
To improve generalization in RL, domain randomization \citep{tobin2017domain, pinto2017asymmetric} is often used to transfer policies from simulation to the real world by utilizing diverse simulated experiences. 
\citet{cobbe2019quantifying, lee2019network} showed that simple augmentation techniques, such as cutout and random convolution, can be useful to improve generalization in RL from images. 
Similarly to us, \citep{laskin2020reinforcement} use data augmentation for RL without any auxiliary loss.
Crucially, most data augmentations techniques in the literature require human knowledge to augment the data according to domain-specific invariances (e.g., through cropping, rotation, or color jittering), and mostly target the learning from image settings.
Nevertheless, heuristics for data augmentation can be formally justified through a causal invariance assumption with respect to certain perturbation on the inputs. 

\fs{Offline learning, distributional shift and causal confusion}
Offline RL and imitation learning methods rely on the availability of informative demonstrations \citep{lange2012batch, li23b, urpi2023efficient, lynch2019play, vlastelica2021neuro}. 
However, in low-data regimes or task-agnostic demonstrations settings, these methods often suffer from distributional shift. 
This shift occurs when the agent induces a state-action distribution which deviates from the original data \citep{ross2011reduction}. 
Several offline learning methods have been proposed for fighting distributional shift, such as by  minimizing deviation from the behavior policy \citep{fujimoto2019off, kumar2019stabilizing, kostrikov2021offline, urpi2021risk} or minimizing risk \citep{vlastelica2021risk}. 
In imitation learning, several works focus on solving the causal confusion problem \citep{de2019causal}, where a policy exploits nuisance correlates in the states for predicting expert actions \citep{wen2020fighting, seo2024regularized}. 
\citet{gupta2023can} propose mitigating spurious correlations in offline RL by upsampling transitions with high epistemic uncertainty in the advantage function.

\fs{Causal Reinforcement Learning}
Detecting causal influence involves causal discovery, which can be pictured as finding the existence of arrows in a causal graph \citep{pearl2009causality}.
While it remains an unsolved task in its broadest sense, there are assumptions that permit discovery in some settings \citep{identif2012, spirtes2001causation}.

Once the existence of an arrow can be detected, its impact needs to be established, for which several measures, such as transfer entropy or information flow, have been proposed \citep{schreiber2000measuring,lizier2012local, ay2008information}. 
In our case, we use conditional mutual information \citep{cover1999elements} as a measure of causal action influence, as proposed by \citet{Seitzer2021CID}.

The intersection of RL and causality has recently been studied to improve interpretability, sample efficiency, and to learn better representations \citep{buesing2018woulda, bareinboim2015bandits, lu2018deconfounding, rezende2020causally}.

\citet{lyle2021resolving} dealt with the problem of causal hypothesis-testing in the online setting via an exploration algorithm,
\citet{ding2023seeing} utilize a model-based counterfactual approach~\citep{pitis2022mocoda} for solving a robust MDP and
\citet{zhang2020learning} investigate how to obtain a reduced causal graph using a block structure.
While \citet{wang2022causal} also leverage influence to learn the causal structure, they use it for state abstraction to improve model generalization. Similarly to us, ~\citep{lu2020sample, ding2022generalizing}, also provide the agent with counterfactuals, but by learning and querying a model for the state transition.
In particular, our work is related to that of \citet{Pitis2020}, which proposes the Local Causal Model framework to generate counterfactual data, and underpins our work. However, \citet{Pitis2020} aim at estimating the entire local causal graph, which is a challenging  problem. In practice, they rely on a heuristic method based on the attention weights of a transformer world model, which does not scale well to high-dimensional environments.
In contrast, our method does not require learning the entire local causal graph, as it assumes that the interactions between entities (except the agent) are sparse enough to be neglected. 
This also implies that the agent is the only entity that can influence the rest of the entities through its actions.
Therefore, this setting is related to the concept of contingency awareness from psychology \citep{contingency1966}, which was interestingly already considered in deep reinforcement learning methods for Atari \citep{song2020, choi2018}. 



\section{Experiments}\label{sec:experiments}
We evaluate \method in two goal-conditioned settings: offline RL and offline self-supervised skill learning. In particular, we are interested in evaluating whether \method \textbf{(i)}~leads to better robustness to extreme distributional shifts, \textbf{(ii)}~enlarges the support of the joint distribution over the state space in low data regimes, and \textbf{(iii)}~works as an independent module combinable with arbitrary learning-based control algorithms. 

The set of benchmarks revolves around the issue of spurious correlations in the state space in manipulation domains, and is built upon the Franka-Kitchen \citep{gupta2019} and the Fetch \citep{plappert2018multi} platforms.
The training data for each task includes spurious correlations that can usually appear during the data collection process, and could distract the policy from learning important features of the state.

\paragraph{Baselines}
We compare \method with \algo{CoDA} \citep{Pitis2020}, a counterfactual data augmentation method, which uses the attention weights of a transformer model to estimate the local causal structure. Given two transitions that share local causal structures, it swaps the connected components to form new transitions. Additionally, we compare with an ablated version of \algo{CoDA}, \algo{CoDA-action}, which only estimates the influences of the action using the transformer weights and thus is a `heuristic'-sibling of our method. For the RL experiments, we also compare it with \algo{RSC} \citep{ding2023seeing}, a framework for robust reinforcement learning that constructs new samples by perturbing the value of the states using an heuristic and learns a structural causal model to predict the next state given the perturbed state. 
As an ablation, we include a baseline without data augmentation (\algo{No-Augm}).
For completeness, we provide comparisons with model-based approaches in \ref{app:mb}.

Given our method and some of the baselines performances depend on an appropriate choice of the parameter $\theta$ to get a classification of influence, we provide a thorough analysis on how this parameter was chosen in \cref{app:theta_study}.
\subsection{Tasks with Spurious Correlations}
Our initial experiments investigate whether \method can increase the generalization capabilities of algorithms when learning from demonstrations that include spurious correlation. Specifically we test whether the trained algorithms are robust to extreme state distributional shifts at test time.

\subsubsection{Goal conditioned offline self-supervised skill learning}

We apply our method to the challenging Franka-Kitchen environment from \citet{gupta2019}.
We make use of the data provided in the D4RL benchmark \citep{Fu2020d4rl}, which consists of a series of teleoperated sequences in which a 7-DoF robot arm manipulates different parts of the environment (e.g., it opens microwave, switches on the stove).
Crucially, all demonstrations are limited to a few manipulation sequences (for example, first opening the microwave, turning on a burner, and finally the light). Thus, the support of the joint distribution over entities in the environment is reduced to only a few combinations. To illustrate this with an example, the light is only on when the microwave is open.
At test time we evaluate the trained agent in unseen configurations, breaking the spurious correlations that exist in the training data.
We hypothesize that \method will create valid counterfactual data such that the downstream learning algorithms would be able to generalize to unseen state configurations.
As a downstream learning algorithm we use LMP \citep{lynch2019play}, an offline goal-conditioned self-supervised learning algorithm, which learns to map similar behaviors (or state-action trajectories) into a latent space from which goal-conditioned plans can be sampled. 
Formally, LMP is a sequence-to-sequence VAE \citep{sohn2015learning, bowman2015generating} autoencoding random experiences extracted from the dataset through a latent space. In our case, we use experiences of fixed window length $\kappa$. Given the inherent temporal abstraction of the algorithm, we generate counterfactuals of fixed length $\kappa >1$ by computing the uncontrollable set for the entire  window as the intersection over all time slices, as in \eqref{eqn:uncontrollable_intersection}.
For specific details on the learning algorithm and the Franka-Kitchen environment, we refer to \ref{app:gc} and \ref{app:kitchen} respectively.


\begin{figure}\centering
    \includegraphics[width=\linewidth]{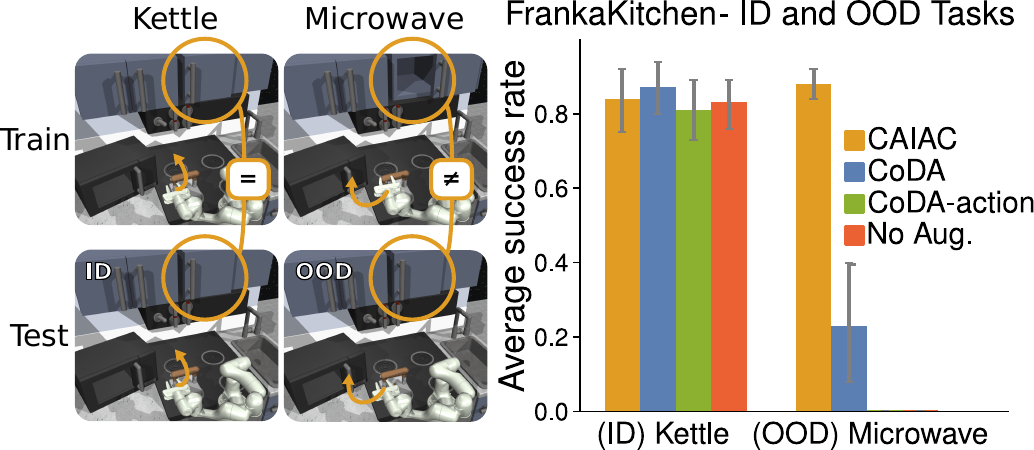}
    \vspace{-0mm}
    \caption{Motivating Franka-Kitchen example. The experimental setup (left) and success rates for in-distribution and out-of-distribution tasks (right). Metrics are averaged over 10 seeds and 10 episodes per task, with 95\% simple bootstrap confidence intervals.}
    \vspace{-0mm}
    \label{fig:kitchen_easy_task_ood}
\end{figure}

\paragraph{Franka-Kitchen:  A Motivating Experiment}\label{exp:toy_kitchen}
Our first experiment is designed to verify claim (i), \ie, that \method enables generalization to unseen configurations over entities.
First, we showcase this in a simple and controlled environment.
Thus, we create a reduced modified dataset from the original D4RL dataset \citep{Fu2020d4rl}, that contains only demonstrations for the microwave task \algo{(mw)} and the kettle \algo{(k)} task. 
During demonstrations for the \algo{(mw)} task, we initialize the cabinet to be always open, whereas for demonstrations for the \algo{(k)} task, it remains closed.
The rest of the objects are set to the default initial configuration (see \ref{app:kitchen}). 
At inference time, we initialize the environment with its default initial configuration (crucially, the cabinet is closed), and we evaluate both tasks (\algo{(k)} and \algo{(mw)}), as shown in \figref{fig:kitchen_easy_task_ood}(left).
Hence, while the \algo{(k)} task was demonstrated for the current configuration (in-distribution, ID), the agent is effectively evaluated on an out-of-distribtion (OOD) configuration for the \algo{(mw)} task.

\looseness -1 We evaluate success rate on both tasks with \method and all baselines, as shown in \figref{fig:kitchen_easy_task_ood}(right). 
All methods are able to solve the \algo{(k)} task, as expected, since it is in-distribution (ID).
However, we observe fundamentally different results for the OOD \algo{(mw)} task.
In principle, \method can detect that the sliding cabinet is never under control of the agent, and will be able to create the relevant counterfactuals to prevent the policy from picking up on the spurious correlation in the data.
Indeed, the performance of \method in the OOD task \algo{(mw)} is not affected, and it is the same as for the ID task. 
On the other hand, the performance of \algo{CoDA} and \algo{CoDA-action} is drastically impaired in the OOD setting.
Despite the simplicity of the setting, the input dimensionality of the problem is high, and the transformer attention weights are not able to recover the correct causal graph.
By picking up on spurious correlations, the attention weights of the transformer estimate low influence from the action to all entities (even the agent), and hence \algo{CoDA-action} creates dynamically-unfeasible counterfactuals which affect performance. 
Since the ratio of observed-counterfactuals data is 1:1 we hypothesize that there is enough in-distribution data to not affect the \algo{(k)} task for \algo{CoDA-action}. 
The local graph induced by \algo{CoDA} has at least as many edges as the one of \algo{CoDA-action}, and hence the probability for creating unfeasible counterfactuals is lower. 
We hypothesize, that despite not learning correct causal influence, it might still provide some samples which benefit the learning algorithm and allow for an average OOD success rate of $0.2$.
We refer the reader to Appendix \ref{app:ablation} for further analysis on the impact of the ratio of observed:counterfactual data for this experiment. 
Finally, as expected, \algo{No Augm.} fails to solve the OOD \algo{(mw)} task.

\paragraph{Franka-Kitchen: All Tasks} \label{exp:full_kitchen}
Having evaluated \method in a controlled setting, we now scale up the problem to the entire Franka-Kitchen D4RL dataset. While in the standard benchmark the agent is required to execute a single fixed sequence of tasks, we train a goal-conditioned agent and evaluate on the full range of tasks, which include the microwave, the kettle, the slider, the hinge cabinet, the light switch and the bottom left burner tasks \cite{mendonca2021discovering}. One task is sampled for each evaluation episode. While alleviating the need for long-horizon planning, this results in a challenging setting, as only a subset of tasks is shown directly from the initial configuration. However, the largest challenge in our evaluation protocol lies in the creation of unobserved state configurations at inference time. While the provided demonstrations always start from the same configuration (e.g., the microwave is always initialized as closed), at inference time, we initialize all non-target entities (with $p=0.5$) to a random state, hence exposing the agent to OOD states.
We expect that agents trained with \method will show improved performance to unseen environment configurations, as those can be synthesized through counterfactual data augmentation.
The results, shown in \tabref{table:all-kitchen}, are consistent with the challenging nature of this benchmark, as the evaluated tasks involve OOD settings in terms of states and actions. Nevertheless, we find that \method is significantly better than baselines in 6/7 tasks, while the last task is unsolved by methods. We hypothesize that the low performance on some of the tasks is due to the absence of robot state and action trajectories in the dataset that show how to solve each of the tasks from the initial robot joint configuration. Hence, even with perfect counterfactual data augmentation these tasks remain challenging. We refer the reader to the Appendix \ref{app:kitchen} for further analysis. 
As observed in the simplified setting, methods relying on heuristic-based causal discovery (\algo{CoDA} and \algo{CoDA-action}) suffer from misestimation of causal influence, and thus from the creation of dynamically-unfeasible training samples. See  \ref{app:test_quality_cf} for a further analysis on the quality of the created counterfactuals and \figref{fig:cai_scores_kitchen} for a visualization of the computed CAI scores per each entity on one of the demonstrations for the Franka-Kitchen dataset.
Finally, without any data augmentation, the learning algorithm \ie \algo{No Augm.} baseline) fails to perform the OOD tasks.

\begin{table}[t]
    \centering
    \caption{
    Average success rates for Franka-Kitchen tasks with OOD initial configurations, computed over 10 seeds and 20 episodes per task with 90\% simple bootstrap confidence intervals.
    \label{table:all-kitchen}}
    \vspace{-2mm}
\scalebox{0.72}{
\begin{tabular}{@{}ccccccc@{}}
    \toprule
    &  \textbf{Algorithm}  & \textbf{\method} & CoDA & CoDA-action & No-Augm.\\ \midrule
    \multirow{3}{*}{\STAB{\rotatebox[origin=c]{90}{\small{}}}} 
    & Kettle &        \textbf{0.81 $\pm$ 0.07}    &         0.18 $\pm$ 0.05            & 0.16 $\pm$ 0.07       & 0.07 $\pm$ 0.06 \\
    & Microwave & \textbf{0.75  $\pm$ 0.09}    &     0.07 $\pm$ 0.05  &     0.0 $\pm$ 0.03      &     0.01 $\pm$ 0.03  \\
    & Bottom-burner & \textbf{0.13  $\pm$ 0.05} &         0.01 $\pm$ 0.03   &            0.0 $\pm$ 0.02 &        0.01 $\pm$ 0.02 \\
    & Slide cabinet &    \textbf{0.14 $\pm$ 0.04} &             0.1 $\pm$ 0.03    &     0.02 $\pm$ 0.02    &  0.07 $\pm$ 0.03 \\
    & Light switch & \textbf{0.01  $\pm$ 0.01} &          0.0 $\pm$ 0.0   &           0.0 $\pm$ 0.0  & 0.00 $\pm$ 0.0\\
    & Hinge cabinet & 0.0  $\pm$ 0.0 &             0.0 $\pm$ 0.0   &          0.0 $\pm$ 0.0  &         0.0 $\pm$ 0.0  \\
   \bottomrule
\end{tabular}
}
\vspace{-5mm}
\end{table}

\begin{figure*}[thb!]
     \centering\vspace{0em}
     \begin{tabular}{ccccc}
        \multicolumn{2}{c}{Fetch-Pick\&Lift with four cubes}
        & \hfill
        &\multicolumn{2}{c}{Fetch-Push with two cubes} \\
            \includegraphics[height=3.1cm]{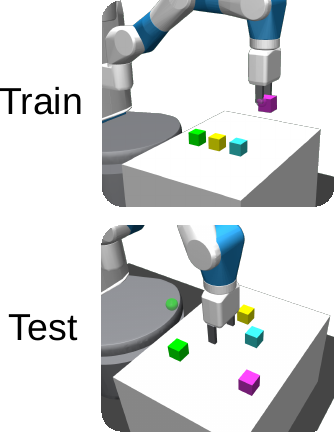}
    &    \includegraphics[height=3.1cm]{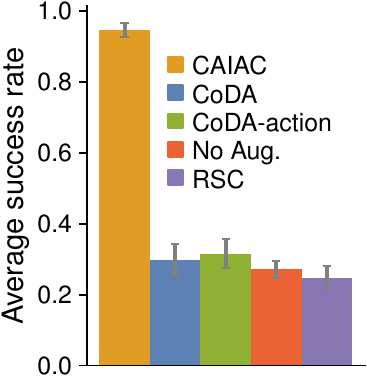}
& \hfill
&\includegraphics[width=0.2\linewidth,trim={0 -4cm 0 0}]{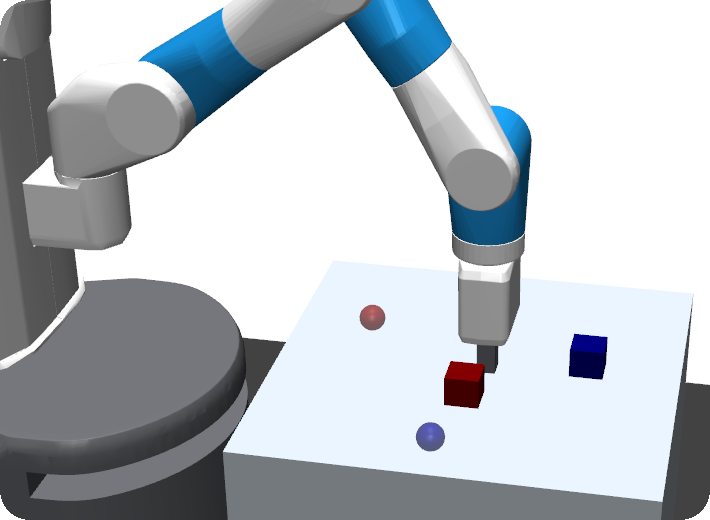}
        &
        \includegraphics[height=3.2cm]{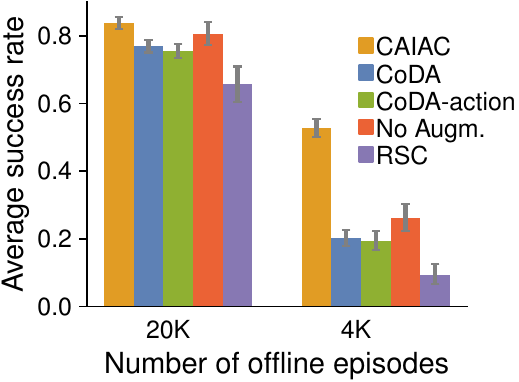}
    \end{tabular}
    \vspace{-0mm}
         \caption{Success rates for Fetch-Pick\&Lift with 4 objects (left) and Fetch-Push with 2 cubes (right). Metrics are averaged over 30 seeds and 50 episodes with 95\% simple bootstrap confidence intervals. }
    \label{fig:fetch}
    \vspace{-0mm}
\end{figure*}

\subsubsection{Goal-conditioned Offline RL}\vspace*{-.0em}
\paragraph{Fetch-Pick\&Lift with 4 cubes}
\label{exp:fetchpicklift}

We additionally test \method on Fetch-Pick\&Lift, a modified version of the Fetch-Pick\&Place environment \citep{plappert2018multi}  were a robot needs to pick and lift a desired cube out of 4 arranged on a table (\figref{fig:fetch} (left)).
For this benchmark, we include spurious correlations in the training data by always arranging the cubes in a line. At test time, the cubes are randomly positioned on the table, evaluating the agent in out of distribution states.
We collect 40k trajectories using an expert policy (50\%) and random policy (50\%) and train an agent offline using TD3+BC \citep{fujimoto2021minimalist}.

Results are shown in \figref{fig:fetch} (left). 
We observe that \method reaches a high success rate by creating relevant counterfactuals that augment the support of the joint distribution over entities and break the spurious correlations in the data. Conversely, the rest of the baselines exhibit subpar performance.
Once again, for \algo{CoDA} and \algo{CoDA-action}, the attention weights of the transformer fail to recover the correct causal graph, resulting in the generation of infeasible samples.  On the other hand, the causally uniformed heuristics used to perturb the states in \algo{RSC}, might break the true cause and effect relationships between state dimensions, leading to performance drop, as reported in \citep{ding2023seeing}.
Moreover, there is no theoretical guarantee that the learned dynamics model, which is regularized for improved generalization, is inherently causal. Consequently, the generated next-state samples may be inaccurate. Further details are given in Appendices \ref{app:td3_bc} and \ref{app:fetchpicklift}.

\subsection{Low Data Regimes}
\subsubsection{Goal-conditioned Offline RL}\vspace*{-.0em}
With this final experiment, our aim is to verify claim (ii), \ie, that \method can enlarge the support of the joint distribution in low data regimes, even when spurious correlations are not necessarily present in the training data, and no distributional shift is injected at test time.
\paragraph{Fetch-Push with 2 cubes}
\label{exp:fetch_push}
We evaluate \method in a Fetch-Push environment \citep{plappert2018multi}, where a robotic arm has to slide two blocks to target locations.
For this experiment we collect 20k trajectories using an expert policy (30\%) and random policy (70\%) and train an agent offline using TD3 \citep{fujimoto2018addressing} in two data regimes: namely 100\% and 20\% of data.

More details are given in Appendices \ref{app:td3}, \ref{app:fetchpush} and \ref{app:theta_study}.
We compare success rates between baselines and \method among different data regimes in \figref{fig:fetch} (right).

In the high data regime, \method and \algo{No Augm.} baseline perform similarly given that there is enough coverage of the state space in the original dataset.
In contrast, in the low data regime \method performs significantly better. Given that the samples in the data, cover sufficient support of the marginal distribution of each entity, \method can substantially increase the support of the joint distribution over entities, leading to higher performance. 
%
Transformer-based methods \algo{CoDA} and \algo{CoDA-action}, and \algo{RSC} create detrimental counterfactuals in all data regimes leading to decreased performance.  To showcase the previous claims, the estimated influence scores for all the methods are visualized in \ref{app:cai_settings}.
We note that, while previous work \citep{Pitis2020} has shown good online performance of \algo{CoDA} in this environment, it resorted to a handcrafted heuristic to decide about influence.

\vspace*{-.0em}
\section{Discussion}\vspace*{-.0em}
While extracting complex behaviors from pre-collected datasets is a promising direction for robotics, data scarcity remains a principal issue in high-dimensional, multi-object settings, due to a combinatorial explosion of possible state configurations which cannot be covered densely by demonstrations. Hence, current learning methods often pick up on spurious correlations and struggle to generalize to unseen configurations.
In this paper, we propose \method as a method for counterfactual data augmentation without the need for additional environment interaction or counterfactual model rollouts, which can be used with any learning algorithm.
By adding an inductive bias on the causal structure of the graph, we circumvent the problem of full causal discovery and reduce it to the computation of an explicit measure of the agent's causal action influence over objects.
Empirically, we show that \method leads to enhanced performance and generalization to unseen configurations, suggesting that further advances in addressing both partial and full causal discovery problems can be substantially beneficial for robot learning.
While our current approach deems action influence to be more important than object-object interaction, in future work, we aim to explore alternative forms of independence to make our approach applicable to a wider range of tasks.
Finally, we would like to further investigate rebalancing the data distribution to counteract data imbalances in the dataset.

\paragraph{Reproducibility Statement}
In order to ensure reproducibility of our results, we make our codebase publicly available at \url{https://sites.google.com/view/caiac}, and provide detailed instructions for training and evaluating the proposed method. Furthermore, we describe algorithms and implementation details in Appendix \ref{ap:full}.
Finally, as our experiments rely on offline datasets, we publish them at the same link.


\section*{Acknowledgements}
The authors thank Cansu Sancaktar and Max Seitzer for their help reviewing the manuscript and the annonymous reviewers for their valuable feedback.
The authors thank the Max Planck ETH Center for Learning Systems for supporting Núria Armengol and Marco Bagatella, and the International Max Planck Research School for Intelligent Systems for supporting Marin Vlastelica.
Georg Martius is a member of the Machine Learning Cluster of Excellence, funded by the Deutsche Forschungsgemeinschaft (DFG, German Research Foundation) under Germany’s Excellence Strategy – EXC number 2064/1 – Project number 390727645.
This work was supported by the ERC - 101045454 REAL-RL.

\section*{Impact Statement}
This paper aims to push the boundaries of generalization in machine learning. Hence, it shares the many societal consequences tied to the field of machine learning, from ethical to environmental consequences.
\bibliography{bibl}
\bibliographystyle{icml2024}
\clearpage

\newpage
\appendix
\onecolumn

\section{Appendix} \label{ap:full}
\subsection{Implementation of downstream learning algorithms}
In this section, we report implementation details concerning the learning algorithms. For a fine-grained description of all hyperparameters, we refer to our codebase at \url{https://sites.google.com/view/caiac}.

\subsubsection{Goal-conditioned offline self-supervised skill learning} \label{app:gc}

For the goal-conditioned self-supervised learning experiments we used LMP \citep{lynch2019play}, a goal-conditioned self-supervised method. It consists of a stochastic sequence encoder, or learned posterior, which maps a sequence $\tau$ to a distribution in latent plan space $q(z | \tau)$, a stochastic encoder or learned goal-conditioned prior $p(z |s, g)$ and a decoder or plan and goal conditioned policy: $\pi(a | z, s,g) $.
The self-supervised goals $g$ are relabeled from achieved goals in the trajectory.For counterfactual samples, trajectories are augmented \textit{before} goal sampling.
The main difference with the original implementation is that the latent goal representation is only added to the prior, but not the decoder.
Additionally, we also implemented KL balancing in the loss term between the learned prior and the posterior: we minimize the KL-loss faster with respect to the prior than the posterior.
Given that the KL-loss is bidirectional, in the beginning of training, we want to avoid regularizing the plans generated by the posterior towards a poorly trained prior. Hence, we use different learning rates, $\alpha = 0.8$ for the prior and $1-\alpha$ for the posterior, similar to \citet{hafner2020mastering}.
These two modifications were also suggested in \citep{rosete2022tacorl}.
Additionally, our decoder was open-loop (instead of close loop): given a sampled latent plan $z$ it decodes the whole trajectory of length  \code{skill length}$ = N$, \ie our decoder is $\pi(\hat{a} | z)$, where $\hat{a} = a_t, ..., a_{t+N}$ is the sequence of decoded actions, instead of $\pi(a | z, s,g)$.
This modification was needed due to the \textit{skewness} of the dataset. Since the demonstrations were provided from an expert agent, given most of the states, the distribution over actions is unimodal: when the robot is close to the microwave, the only sequence of actions in the dataset is the one that opens the microwave. Hence, a close-loop decoder would learn to ignore the latent plan, and only rely on the state. To solve this issue, we make the decoder open-loop.

\subsubsection{Goal-conditioned Offline RL: TD3}
\label{app:td3}
We implement the TD3 algorithm \citep{fujimoto2018addressing} with HER \citep{andrychowicz2017hindsight}. Unless specified differently, the hyperparameters used were the ones from the original TD3 implementation.
We use HER \citep{andrychowicz2017hindsight} to relabel the goals for real data, with a \code{future} relabeling strategy with $p=0.5$, where the time points were sampled from a geometric distribution with $p_{geom}=0.2$.
For the counterfactual data we relabel the goals with $p=0.5$ random sampling from the achieved goals in the buffer of counterfactual samples. In the experiments, we realized that the relabeling strategy had an impact on the performance of the downstream agent. To disentangle the impact of the relabeling strategy from the impact of the counterfactual data generation and to ensure a fair comparison, we also relabeled the same percentage of goals (\ie $p=0.25$) with \code{random} strategy for the \code{No Augm.} baseline.
We train each method for 1.2M gradient steps, although all methods reach convergence after 600k gradient steps.
For all baselines, the percentage of counterfactuals in each batch is set to $0.5$.

\subsubsection{Goal-conditioned Offline RL: TD3+BC}
\label{app:td3_bc}
We implement the TD3+BC algorithm \citep{fujimoto2021minimalist} with HER \citep{andrychowicz2017hindsight} where a weighted behavior cloning loss is added to the policy update. After tuning, we used $\alpha_{BC}=2.5$  ($\alpha_{BC} \to 1$ recovers Behavior Cloning, while $\alpha_{BC} \to 0$ recovers RL). Unless specified differently, the rest of hyperparameters used were the ones from the original TD3 implementation.
We use HER \citep{andrychowicz2017hindsight} to relabel the goals for real data, with a \code{future} relabeling strategy with $p=0.5$, where the time points were sampled from a geometric distribution with $p_{geom}=0.2$ and 

For the counterfactual data we relabel the goals with $p=1$ with \code{future} strategy \textit{after} augmenting the samples. For the \code{No Augm.} we also relabeled goals with \code{random} strategy with $p=0.25$.
We train each method for 120k gradient steps, although all methods reach convergence after 90k gradient steps.
For all baselines, the percentage of counterfactuals in each batch is set to $0.9$ unless specified.

\subsection{Experimental details}
\subsubsection{Franka-Kitchen} \label{app:kitchen}
We use the kitchen environment from the D4RL benchmark \citep{Fu2020d4rl} which was originally published by \citet{gupta2019}. The D4RL dataset contains different dataset versions: \code{kitchen-complete, kitchen-partial, kitchen-mixed}, which contain 3690, 136950 and 136950 samples respectively, making up to approximately 14 demonstrations for \code{kitchen-complete} and 400 demonstrations for each \code{kitchen-partial} and \code{kitchen-mixed}.
The simulation starts with all of the joint position actuators of the Franka robot set to zero. The doors of the microwave and cabinets are closed, the burners turned off, and the light switch also off. The kettle will be placed in the bottom left burner.
The observation are 51-dimensional, containing the joint positions of the robot (9 dim), the positions of the all the kitchen items (21 dim) and the goal positions of all the items (21 dim).
The length of the episode is 280 steps, but the episode will finish earlier if the task is completed. The task is only considered solved when all the objects are within a norm threshold of 0.3 with respect to the goal configuration. 
While in the standard benchmark the agent is required to execute a single
fixed sequence of tasks, we train a goal-conditioned agent, and evaluate on one task per each evaluation episode. For the Franka-Kitchen motivating example (see \ref{exp:toy_kitchen}) we query for either the kettle or the microwave task, in the Franka-Kitchen: All tasks (see \ref{exp:full_kitchen}) we query for  the full range of tasks, which include the microwave, the kettle, the slider, the hinge cabinet, the light switch and the bottom
left burner tasks. While alleviating the need for
long-horizon planning, this results in a challenging setting, as only a subset of tasks is shown directly from the initial configuration. 
Specifically out of the 1200 demonstrations in the dataset, containing different task sequences, only 3 objects are shown to be manipulated from the initial robot configuration: 60\% of the trajectories solve the microwave task first, 30\% show the kettle task first and 10\% show the bottom burner first. This aligns with the relative performance achieved for those tasks. For the 3 remaining tasks, namely the slide cabinet, the light and the hinge cabinet, there is no demonstration shown directly from the initial configuration and hence the low performance.


\textbf{Franka-Kitchen: Motivating Experiment} For the first experiment (see Subsection \ref{exp:toy_kitchen}), we modify the dataset  version \code{kitchen-mixed} to only contain $\sim$ 50 demonstrations of length $\sim$ 40 timesteps for each \code{(mw)} and \code{(k)} task. 
During demonstrations for the \code{(mw)} task, we initialize the cabinet to be always open, whereas for demonstrations for the \code{(k)} task, it remains closed. The rest of the objects are set to the default initial configuration. The goal configuration for all the objects was set to their initial configuration (as defined above), except for the microwave or the kettle, which were set to the default goal configuration when querying for the \code{(mw)} and \code{(k)} tasks respectively.

\textbf{Franka-Kitchen: All Tasks} For the second experiment (see Subsection \ref{exp:full_kitchen}) we merge the 3 provided datasets \code{kitchen-complete, kitchen-partial, kitchen-mixed}.
For this experiment, each object (except the one related to the task at hand to ensure non-trivial completion), was randomly initialized with $p=0.5$, otherwise it was initialized to the default initial configuration (as defined above). We then modify the desired goal to match the initial configuration for all non-target entities.

\subsubsection{Fetch-Push with 2 cubes}\label{app:fetchpush}
Expert data for the experiment in Subsection \ref{exp:fetch_push} includes 6000 episodes collected by an agent trained online using TD3 and HER up to approximately 95\% success rate.
We additionally collect 14000 episodes with a random agent, which make up for the random dataset.
This sums up to a total of 20000 episodes (each of length 100 timesteps), with 30\% expert data and 70\% random data.
Initial positions and goal positions of the cubes are sampled randomly on the table, whereas the robot is initialized in the center of the table with some additional initial random noise . 
The rewards are sparse, giving a reward of $-1$ for all timesteps, except a reward of $0$ when the position of each of the 2 blocks are within a $2-$norm threshold of $0.05$. The observation space is 34-dimensional, containing the position and velocity of the end effector (6dim), of the gripper (4dim) and the object pose, linear and rotational velocities of the objects (12dim each).
In contrast to the original \code{Fetch-Push-v1} \citep{plappert2018multi} environment and similarly to \citet{Pitis2020} we do not include parts of the state space accounting for relative position or velocities of the object with respect to the gripper, which would entangle the two. The goal is 6-dimensional encoding the position for each of the objects. The action space is 4-dimensional encoding for the end-effector position and griper state.
At test time, we count the episode as successful upon reaching the goal configuration (\ie, observing a non-negative reward).

\subsubsection{Fetch-Pick\&Lift with 4 cubes}
\label{app:fetchpicklift}
Expert data for the experiment in Subsection \ref{exp:fetchpicklift} includes 20000 episodes collected by an agent trained online using TD3 and HER up to approximately 95\% success rate.
We additionally collect 20000 episodes with a random agent, which make up for the random dataset.
This sums up to a total of 40000 episodes (each of length 50 timesteps), with same percentage of expert and random data.
During data collection, we initialise all the 4 cubes aligned (\ie all having the same x-axis position). In each episode, we initialise this value from a categorical distribution with 5 options.
During testing, the initialization process is the same, but it is done independently for each entity. As a result, the cubes are no longer aligned.
The robot is initialized in the center of the table with some additional initial random noise . 
The rewards are sparse, giving a reward of $-1$ for all timesteps, except a reward of $0$ when the position of each all the cubes are within a $2-$norm threshold of $0.05$. The observation space is 58-dimensional, containing the position and velocity of the end effector (6-dimensional), of the gripper (4-dimensional) and the object pose, linear and rotational velocities of the objects (12 dimensions each).
In contrast to the original \code{Fetch-Push-v1} \citep{plappert2018multi} environment and similarly to \citet{Pitis2020} we do not include parts of the state space accounting for relative position or velocities of the object with respect to the gripper, which would entangle the two. The goal is 12-dimensional encoding the position for each of the objects. The action space is 4-dimensional encoding for the end-effector position and griper state.
At test time, we count the episode as successful upon reaching the goal configuration (\ie, observing a non-negative reward).

\subsection{Ablation: Ratio of observed-to-counterfactual Data} \label{app:ablation} 

\begin{figure}\centering
    \includegraphics[height=4.0cm]{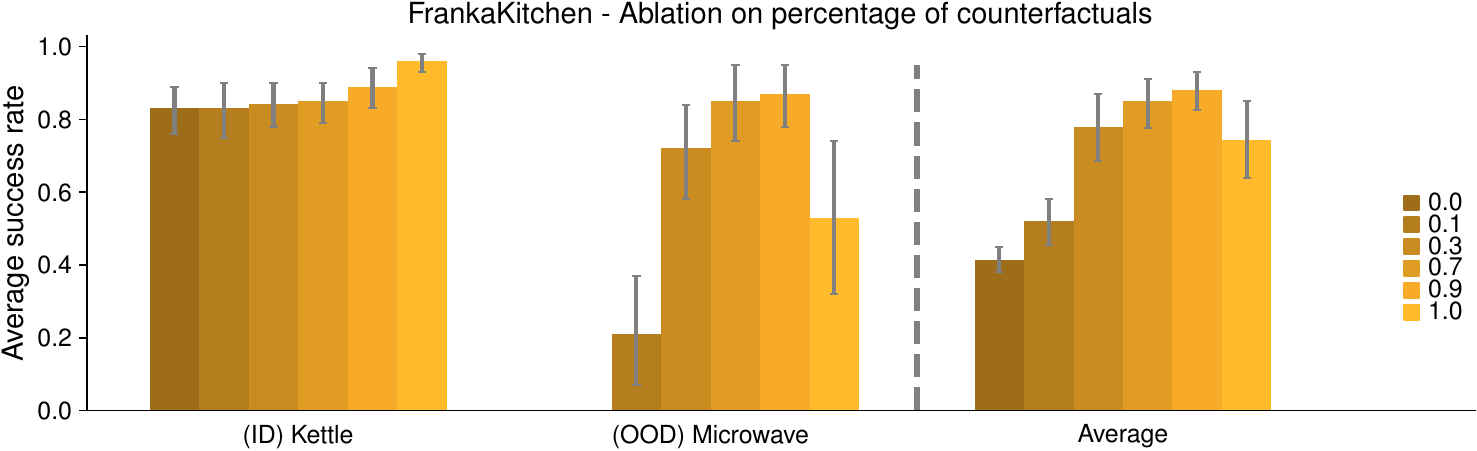}
    \caption{Performance of \method on motivating Franka-Kitchen example when controlling the percentage of counterfactual samples in each batch. Metrics are averaged over 10 seeds and 10 episodes per task, with 95\% simple bootstrap confidence intervals.}
    \label{fig:ablation}
\end{figure}

In this section, we study the effect of the ratio of observed-to-counterfactual data generated with \method, by evaluating downstream performance on the Franka-Kitchen motivating example, as presented in \ref{exp:toy_kitchen}. Empirical results for this ablation are shown in \figref{fig:ablation}. 
As expected, we observe that the ratio of counterfactuals does not have any significant impact on the success rate on the \code{(k)} task. This is because the task is evaluated in distribution, and hence the downstream learning algorithm does not require observing counterfactual experience (but still does not suffer from it).
For the OOD \code{(mw)} task we see that increasing the number of counterfactuals up to a 0.9 ratio has a positive effect in performance, leading the agent to generalize better to the OOD distribution. However, when the ratio is increased up to 1, we only use synthesized counterfactual data. We observe a decrease in performance with high variance among training seeds. This hard ablation shows the need for real data during training to avoid induced selection bias, as also observed in \citep{Pitis2020}.
With a ratio 0.0, we recover the performance of the No Augm. baseline.

\subsection{Details on Influence Detection Evaluation}
\label{app:cai_settings} 
To detect causal action influence we use CAI, as described in \ref{sec:cai}.
\begin{equation}
    C^j(s) := I(S'_j; A \mid S=s) = \mathbb E_{a\sim \pi} \big[ D_{KL} \big( P_{S'_j\mid s,a}   \;\big|\big| \; P_{S'_j\mid s} \big) \big].
\end{equation}
This requires learning the transition model $P_{S'_j\mid s,a}$.

In the case of robotic manipulation environments \textit{physical contact} is not needed for causal action influence as long as the agent can change the object pose, even if indirectly,  in a single simulation step.

\textbf{World model training}
For the Franka-Kitchen experiments all models were trained to predict the full state of the environment.
For increased performance in the Fetch-Push task, all models were trained to predict the next position of the end effector of the agent gripper and of the objects (3 dimensions each). 
For \method the transition model $P_{S'_j\mid s,a}$ is modeled as a Gaussian neural network (predicting mean and variance) that is fitted to the training data $\mathcal{D}$ using negative log likelihood. We used a simple multi-layer perceptron (MLP) with two separate output layers for mean and variance. To constrain the variance to positive range, the variance output of the MLP is processed by a softplus function (given by $\log(1 + exp(x)))$, and a small positive constant of  $10^{-8}$ was added to prevent instabilities near zero. We also clip the variance to a maximum value of 200. For weight initialization, orthogonal initialization is used. 
For the Franka-Kitchen we use larger MLPS, with 3 layers for the simplified and 4 layers for the full experiment, each with 256 units and a learning rate of $8e^{-4}$.


For \code{CODA} and \code{CODA-action} we use a self-implementation of the transformer model.
We use a model with 3 layers and 4 attention heads for the Fetch-Push task and 5 layers and 4 heads for all the Franka-Kitchen tasks, with an embedding space and output space of 128 dimensions each. We also used a learning rate of $8e^{-4}$.

All models were trained for 100k gradient steps, and tested to reach low MSE error for the predictions in the validation set (train-validation split of 0.9-0.1).
We trained all models using the Adam optimizer \cite{kingma2014adam}, with default hyperparameters.

In general, the models were trained using the same data as for the downstream task for all experiments.
However, for the Franka-Kitchen task, we add some additional collected data on the environment when acting with random actions. The reason is that, in order to compute CAI, we query the model on randomly sampled actions from the action space. Due to the expert nature of the kitchen dataset comes from an informed agent, the original dataset might lack random samples and hence we would query the model OOD when computing CAI.
For both experiments in the Franka-Kitchen simplified experiment we added 1x the original dataset of random data. Further experiments on the impact of the amount of random data could be beneficial. This was not needed for the Fetch-Push task since the dataset already contains random action.
We note that this additional data is also provided for training all transformer-based baselines.

\begin{figure}\centering
    \begin{minipage}{0.35\linewidth}
    \vspace{-9mm}
    \includegraphics[width=\linewidth]{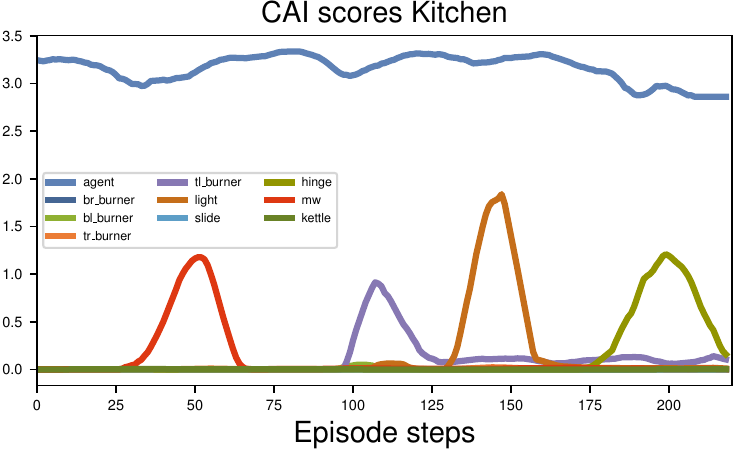}
    \caption{Computed CAI scores per each entity on one of the demonstrations for the Franka-Kitchen dataset. We can observe how the influence of the agent's actions over objects changes over time. First influencing the microwave (mw), then the top-left bottom burner (tl\_burner), then the light switch (light) and finally the hinge cabinet (hinge). We selected a threshold of $\theta = 0.3$.}
    \label{fig:cai_scores_kitchen}        
    \end{minipage}
    \hfill
    \begin{minipage}{0.6\linewidth}
    \includegraphics[height=3.5cm]{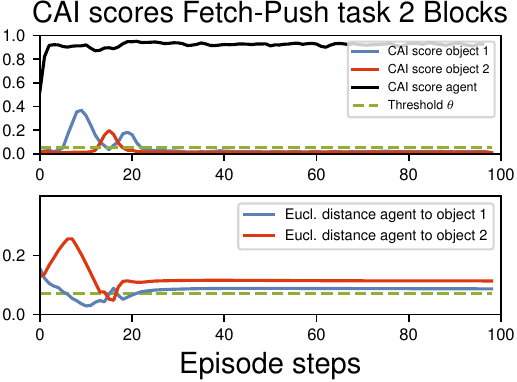}
    \includegraphics[height=3.5cm]{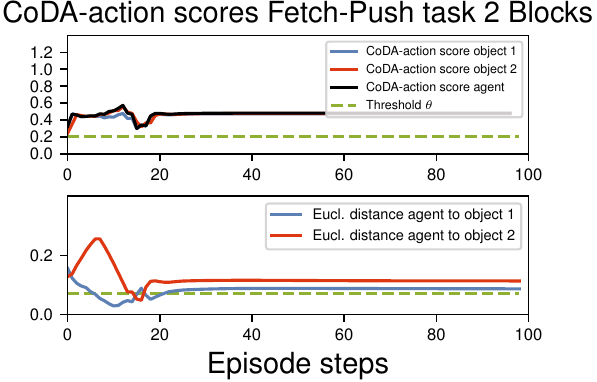}
    \caption{Top left: Computed CAI scores per object on one of the expert demonstrations for the Fetch-Push dataset. We can observe how the influence of the agent's actions over objects changes over time. First pushing object 1, then object 2 and object 1 again. In green we show the optimized threshold $\theta=0.05$ for this task.
    Bottom left and right: We show distance from the robot end-effector to each of the objects as a domain knowledge heuristic for action influence. 
    In green we show the heuristic distance of 7cm that we use as a threshold to consider the agent can influence the object within the next timestep. Top right: Computed CoDA-action scores on the same episode as above.  In green we show the optimized threshold $\theta=0.2$ for this task. } 
    \label{fig:cai_scores_fetchpush}
    \end{minipage}
\end{figure}

\textbf{CAI scores}
In practice, we compute the CAI scores using the estimator:
\begin{equation}
    C^j(s) = \frac{1}{K} \sum_{i=1}^{K} [ D_{KL} \big ( p(s'_j \mid s,a^{(i)} ||  \frac{1}{K} \sum_{k=1}^{K}  p(s'_j \mid s,a^{(k)}               \big)           ]
\end{equation}
with $a \sim \pi$, where $\pi(A) := \mathcal{U}(\mathcal A)$ (\ie a uniform distribution over the action space) and with $K=64$ actions. We refer the reader to \citep{Seitzer2021CID} for more details. In \figref{fig:cai_scores_kitchen} we show the computed CAI scores over a demonstration in the Franka-Kitchen and in \figref{fig:cai_scores_fetchpush} ( left) over an episode for the Fetch-Push task.
Given the scores we need a threshold $\theta$ to get a classification of control (see Equation \ref{eq:uncontr_set}). 
For Fetch-Push we set $\theta = 0.05$, for Fetch-Pick\&Lift we set $\theta = 2.0$ and for Franka-Kitchen $\theta = 0.2$ . See section \ref{app:theta_study} for a thorough analysis on the impact of the influence threshold $\theta$ and how the parameter was chosen for each of the methods.

\begin{figure}\centering
    \includegraphics[width=\linewidth]{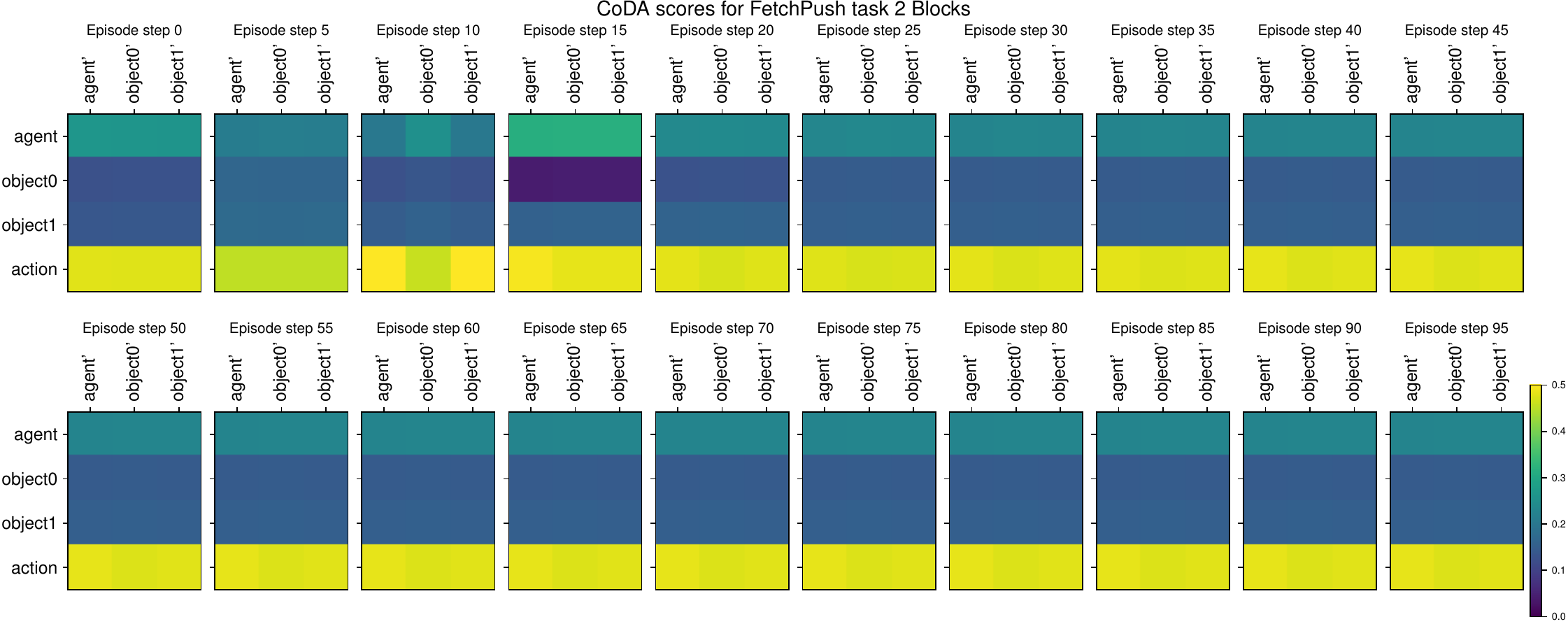}
    \caption{Computed CoDA scores on the same episode as in \figref{fig:cai_scores_fetchpush}. We show snapshots of the attention weights of the transformer every 5 time steps as computed by the CoDA algorithm. The element ${i,j}$ in the matrix shows the attention (or influence) $s_i \rightarrow s'_j$. We see how the transformer completely fails on discovering the full causal graph, being even unable to recover influence along the diagonal, i.e., that an entity state at time t influences the entity state at time t+1.} 
    \label{fig:coda_scores_fetchpush}
\end{figure}

\textbf{Transformer scores}
To compute causal influence, baselines use the attention weights of a Transformer, where the score is computed as follows. Letting $A_i$ denote the attention matrix of the i’th of N layers, the total attention matrix is computed as $\prod_{i=1}^N A_i$. For \code{CoDA} the score is computed by checking the corresponding row $i$ and column $j$ for the check $s_i \rightarrow s'_j$, whereas for \code{CoDA-action} we restrict ourselves to  the row corresponding to the input position of the action component, and the output position of the object component. Our implementation follows \citet{Pitis2020}, to which we refer for more details.
In \figref{fig:cai_scores_fetchpush} (right) and \figref{fig:coda_scores_fetchpush}, we show respectively the computed CoDA-action and CoDA scores over an episode for the Fetch-Push task.
Given the scores we need a threshold $\theta$ to get a classification of control. 
For Fetch-Push we set $\theta = 0.2$ for \code{CODA} and \code{CODA-action}, for Fetch-Pick\&Lift we set $\theta = 0.2$ for \code{CODA-action} and $\theta = 0.15$ for \code{CODA}. For Franka-Kitchen we set $\theta = 0.3$ for all methods. See section \ref{app:theta_study} for a thorough analysis on the impact of the influence threshold $\theta$ and how the parameter was chosen for each of the methods.

\subsection{Analysis on Influence Threshold}
\label{app:theta_study} 
\begin{figure}\centering
    \includegraphics[height=4.8cm]{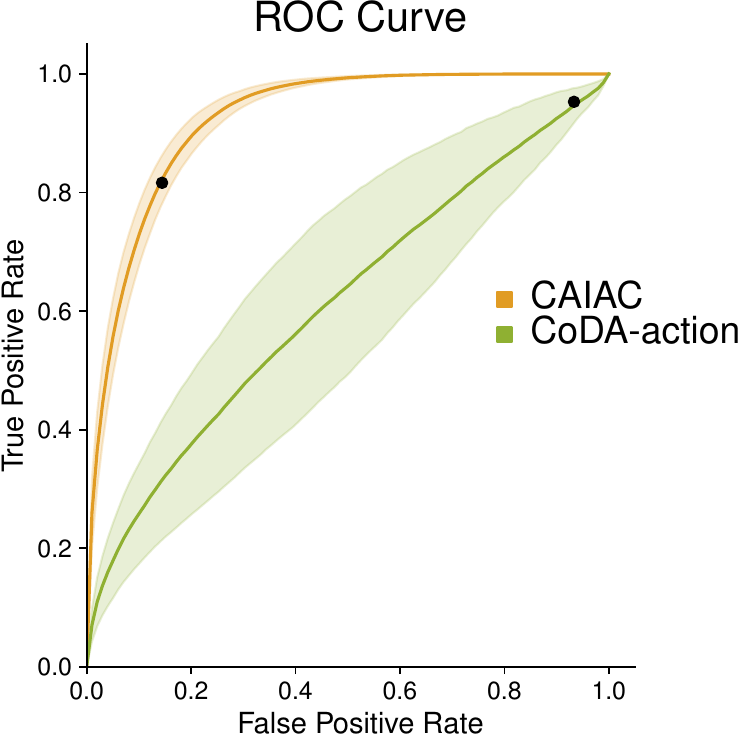}
    \caption{ROC curves for CAIAC and \code{CoDA-action} (averaged over 10 trained world models and 1 standard deviation shaded). We show measures true and false positive rates (TPR and FPR) while sweeping the influence threshold $\theta$. In black we show the corresponding TPR and FPR for the optimal $\theta$ for both methods. See also Figs.~\ref{fig:roc_curves_caiac_all}, \ref{fig:roc_curves_mask_all}.}
    \label{fig:roc_curves_all}
\end{figure}

\begin{figure}\centering
    \includegraphics[width=\linewidth]{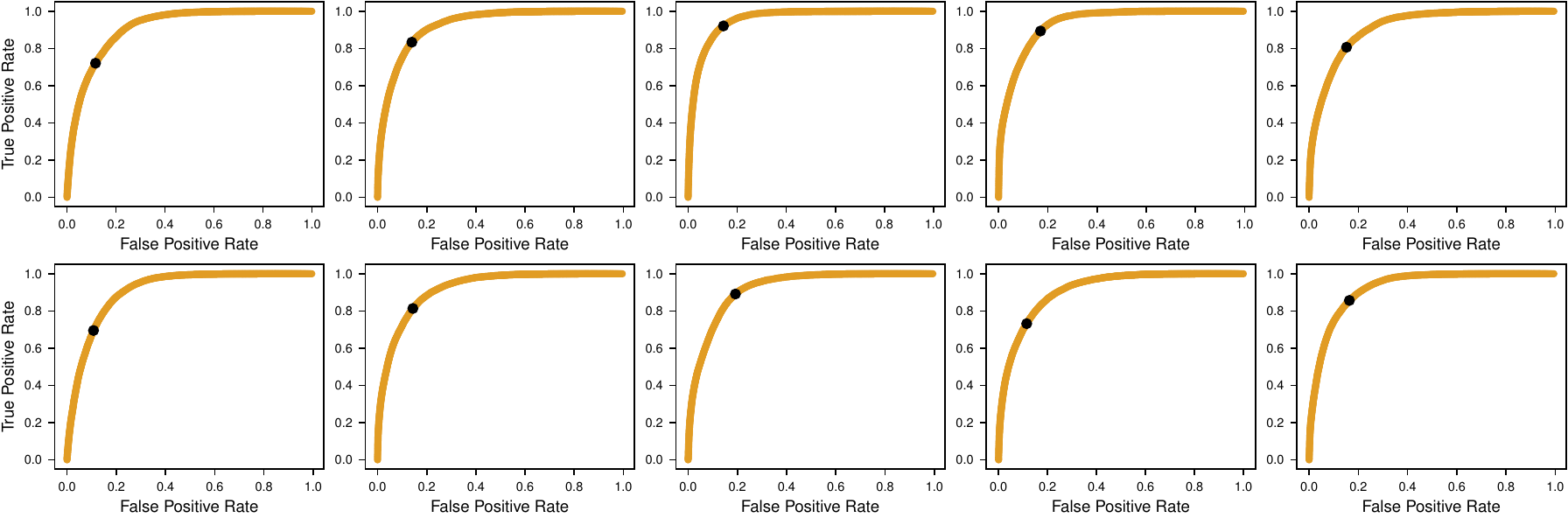}
    \caption{ROC curves for all the 10 trained world using CAIAC. In black we show the corresponding true positive and false positive rate for the optimal threshold $\theta=0.05$.  }
    \label{fig:roc_curves_caiac_all}
\end{figure}

\begin{figure}\centering
    \includegraphics[width=\linewidth]{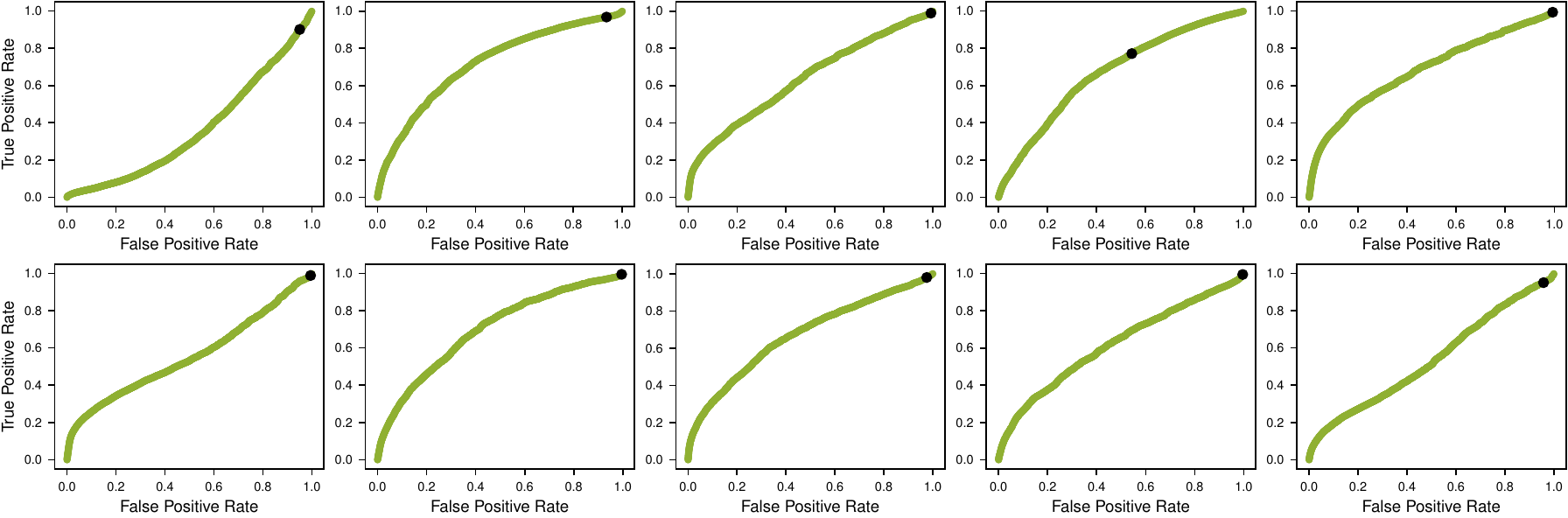}
    \caption{ROC curves for all the 10 trained world using \code{CoDA-action}. In black we show the corresponding true positive and false positive rate for the optimal threshold $\theta=0.2$.}
    \label{fig:roc_curves_mask_all}
\end{figure}

To get a classification of control, we optimize the value for the threshold $\theta$ for all methods.
We train 10 different world models for each method and we run a grid search over the parameter $\theta$. We run 3 seeds for each of the 10 models and we picked the value for $\theta$ that optimizes the downstream task average  performance among the 10 models and the 3 seeds. 
In Figure \figref{fig:roc_curves_all} we plot the ROC curves for correct action influence detection for both CAIAC and \code{CoDA} for the Fetch-Push task. In such an environment, one can specify a heuristic of influence using domain knowledge, namely the agent does not have influence on the object if 7cm apart. An accurate model generates an Area Under the Curve (AUC) close to 1, while a random model stays along the diagonal. In \figref{fig:roc_curves_all} we can observe that the attention weights of the transformer world model are not accurate for detecting influence. Additionally, there is a high variability on the different trained transformers (see also \figref{fig:roc_curves_mask_all}) , making it hard to optimize for the threshold $\theta$ for this type of model architecture. In contrast, we see that ROC curves for CAI have an AUC $\approx 0.9$ and hence it is an accurate measure for predicting influence in the Fetch-Push environment. Additionally, given its low variance across training seeds (see also \figref{fig:roc_curves_caiac_all}), same thresholds reach the same TPR/FPR across models, making it easy to optimize for $\theta$.

\subsection{Computational Demands}
CAIAC relies on computing the CAI measure for data augmentation. In turn, CAI can be evaluated for all entities at once, through $k$ forward passes for the $k$ counterfactual actions, which are performed in a batch-wise fashion.
$k$ is a constant factor, and does not scale with the number of entities.
Methods relying on a transformer world model, like CoDA and CoDA-action only need one forward pass (which internally has quadratic cost in the number of entities due to cross attention). However, CoDA also needs to compute the connected components from the adjacency matrix, which has a quadratic cost.
For relatively few entities, as is common in the robotic manipulation environments, the computational overhead is relatively small.  Table \ref{tab:runtime} reports an evaluation for the high data regime in the Fetch-Push environment, in which each method was timed while computing influence on all 2M datapoints.
The algorithms were benchmarked on a 12-core Intel i7 CPU.
We note that counterfactuals could be generated in parallel to the learning algorithm and hence not significantly impact runtime of the algorithm. Furthermore, in our offline setting, counterfactuals can be fully precomputed.

\begin{table*}[hb!]
    \centering
    \caption{
    Computational demands for computing counterfactuals for the different algorithms. Runtime was benchmarked on a 12-core Intel i7 CPU.
   }
    \vspace{-.5em}
\begin{tabular}{@{}ccccccc@{}}
    \toprule
    &  \textbf{}  & \method & CoDA & CoDA-action\\ \midrule
    \multirow{3}{*}{\STAB{\rotatebox[origin=c]{90}{\small{}}}} 
    & Runtime (min) &        $\sim$13    &         $\sim$10           & $\sim$1      \\
   \bottomrule
\end{tabular}
\label{tab:runtime}
\end{table*}

\subsection{Analysis on the Quality of Created Counterfactuals}\label{app:test_quality_cf}
We provide an analysis on the created counterfactuals using \method and several baselines, which investigates whether each method (i) creates \textit{feasible} samples (\ie in accordance to the true transition kernel of the environment) and (ii) increases the support of the joint state space distribution in the training data.

\paragraph{Feasibility}
The procedure estimating whether the augmentation procedure is valid is as follows. A set of $N$ trajectories  $(s_t,a_t, \dots, a_{t+\tau -1}, s_{t+\tau})$ sampled from the Franka-Kitchen dataset is considered. Counterfactuals trajectories $(\tilde s_t, a_t, \dots, a_{t+\tau -1},\tilde s_{t+\tau})$ are created for each original trajectory using each method.
We then leverage access to the environment's simulator, reset it to the initial counterfactual state $\tilde s_t$, and act on the environment by apply actions $a_t, \dots, a_{t+\tau -1}$.
If the counterfactual is valid, the resulting state of the simulation $s^{SIM}_{t+\tau}$ should coincide with the counterfactual $\tilde s_{t+\tau}$.
To account for non-determinism in the simulator, each action sequence is simulated $K=50$ times, which results in a set of $K$ final states $\mathcal S' = \{\bar s^k_{t+\tau}\}_{k=1}^K$.
A multivariate Gaussian distribution is then fit to the samples from $\mathcal S'$ via Maximum Likelihood Estimation.

This allows the computation of the likelihood of the final counterfactual state $\tilde s_{t+\tau}$ under the Gaussian distribution for each method, and for each of the $N$ initial trajectories. 
\figref{fig:quality_cf} presents the density of log-likelihoods for each of the methods, approximated with a Gaussian KDE.
We observe that \method 's augmented data have high likelihood under the distribution of final states returned by the simulator, and hence are mostly valid.
In contrast, while some counterfactual trajectories generated by other methods are also viable, most of their samples are associated to low log-likelihoods, which suggests a violation of the environment's dynamics.

\paragraph{Increased support}
\figref{fig:quality_cf} (right) shows 1000 randomly selected samples from the Franka-Kitchen dataset, as well as one counterfactual (created using \method) for each. The visualization employs t-SNE for a 2D representation of the high-dimensional state space. 
We observe how the augmented samples cover a larger space than the observed data, suggesting that the support of the joint training distribution over entities is improved. 
We also provide numerical evidence by comparing the support of the empirical distributions.
For each sample augmented with all methods, we first bin each of the object states into 2 categories. We then compute the support of the resulting categorical distribution with $2^N$ possible categories, where $N$ is the number of objects.
Table \ref{table:support_num} reports ratios between the estimated support and the theoretical maximum.

\begin{table*}[hb!]
    \centering
    \caption{
    Support of the empirical joint state space distribution using different augmentation methods (results computed over 1000 augmented samples).Values represent ratios between the estimated support and the theoretical maximum.
   }
    \vspace{-.5em}
\begin{tabular}{@{}ccccccc@{}}
    \toprule
    &  \textbf{}  & \method & CoDA & CoDA-action & No Augm.\\ \midrule
    \multirow{3}{*}{\STAB{\rotatebox[origin=c]{90}{\small{}}}} 
    & Support &     0.52       &         0.48          &  0.42  & 0.3     \\
   \bottomrule
\end{tabular}
\label{table:support_num}
\end{table*}

In \ref{table:support_num} we show how \method increases the support of the original state space distribution by almost a factor of 2 . However, we can see that rest of the baselines also increase it significantly.
Importantly, this numbers do not inform on whether these counterfactuals are actually valid, \ie follow the true transition kernel of the environment, which was already investigated in the above Feasibility paragraph.




\section{Implementation Details for Baselines}
In the following we provide implementation details for \code{RSC} and additional model-based baselines.
Details regarding \method and \code{CoDA} and \code{CoDA-action} baselines were already provided in the respective sections above.
\label{rsc}

\subsection{RSC \citep{ding2023seeing}}
\label{app:rsc}
Due to lack of public code available by the time of the submission, we resort to reimplementing RSC \citep{ding2023seeing}.
The implementation follows the guidelines described in the original paper, with a few differences.
In the original method, the dynamics model leverages a learned causal graph as a gating mechanism to achieve better generalization to perturbed state and action pairs; training is regularized to encourage sparsity in this graph.
In our case, we instead leverage a transformer dynamics model, in which sparsity is granted by the softmax in its attention mechanism.
Additionally, while the dynamics model in RSC is trained to predict both rewards and next states, in our case predicting the next state is sufficient, as the ground truth reward can be computed with the relabeling function available in goal-conditioned settings.

A final difference that ensures a fair comparison is in the perturbation phase.
The original description of RSC is not object-centric, and thus only perturbs a single dimension of the state space.
In our experiments, we ensure that RSC also leverages the known decomposition of the state space in objects, and thus augment the state of a single object in a consistent manner.

Finally, a naive maximization of the heuristic (Equation 7 in \citet{ding2023seeing})over the entire dataset $\mathcal{D}$ and $K$ objects would require evaluating $K|\mathcal{D}|^2$ distances. Due to compute constraints, we optimize it by subsampling 1000 candidate solutions, which we found to still approximate the maximum reasonably well.



\subsection{Model based baselines}
\label{app:mb}
We present a performance comparison between \method and an MBPO-style causal-unaware model-based approach, which we call \code{MBPO} for simplicity.

\code{MBPO} leverages a dynamical model trained from data to generate on-policy imagined N-step transition rollouts.
Despite the ease of data generation, these methods often suffer from bias of model-generated data, which leads to approximation errors that compound with increasing the horizon (N).

In our spurious correlation experiments, we would argue that it is hard for the model to accurately autoregressively predict long horizons trajectories, that would lead the agent to OOD states useful to solve the task at hand. Additionally, if that would be the case, then a causal-agnostic model would be queried OOD suffering leading to poor generalization. Work on the direction for task independent state abstraction tackles this issue \citep{wang2022causal}.

\begin{figure*}[thb!]
     \centering\vspace{0em}
     \begin{tabular}{ccccc}
        \multicolumn{2}{c}{Fetch-Pick\&Lift with four cubes}
        & \hfill
        &\multicolumn{2}{c}{Fetch-Push with two cubes} \\
            \includegraphics[height=3.1cm]{figures/fpp/env_fpp.pdf}
    &    \includegraphics[height=3.1cm]{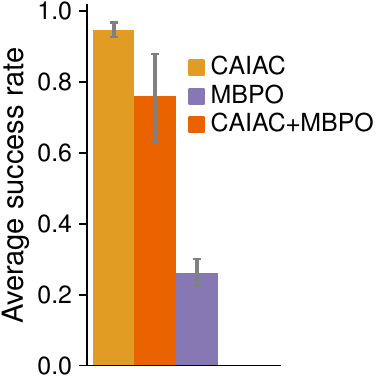}
& \hfill
&\includegraphics[width=0.2\linewidth,trim={0 -4cm 0 0}]{figures/fetch/fetch_push_round.png}
        &
        \includegraphics[height=3.2cm]{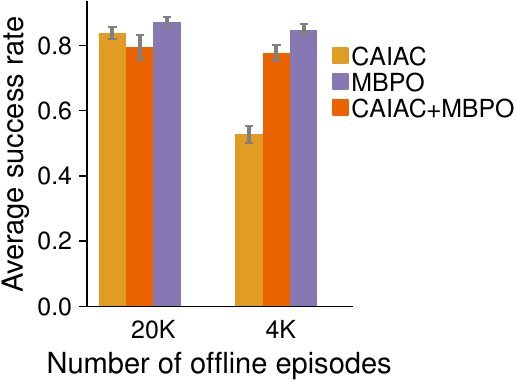}
    \end{tabular}
    \vspace{-0mm}
         \caption{Success rates for CAIAC and model-based approaches in Fetch-Pick\&Lift with 4 objects (left) and Fetch-Push with 2 cubes (right). Metrics are averaged over 30 seeds and 50 episodes with 95\% simple bootstrap confidence intervals. }
    \label{fig:dyna_baselines}
    \vspace{-0mm}
\end{figure*}

We show the results for the reinforcement learning experiments (namely FetchPush and Fetch-Pick\&Lift) in \figref{fig:dyna_baselines}.

For the spurious correlation environment \figref{fig:dyna_baselines} (left), we indeed confirm our hypothesis, were \code{MBPO} is unable to create samples that break the spurious correlation in the data, and hence leads to poor performance compared to \method.

However, in the low data regime environments \figref{fig:dyna_baselines} (right), we observe how \code{MBPO} thrives and outperforms \method.
We hypothesize that in this environment, the trained model is accurate enough even in low data regimes, and hence can generate useful trajectory rollouts that increase the amount of data substantially. Since spurious correlations are not present in the data, the model doesn't need to be queried OOD to create meaningful samples.

We finally decide to leverage the strengths of both approaches and propose \code{CAIAC+MBPO}. With this approach, we train a dynamics model both \textit{on the original AND \method 's augmented data}. Notably, the model is trained on perturbed yet dynamically feasible samples, potentially leading it to better generalization capabilities in OOD states. 
Finally, we we train our agent with \code{MBPO} using both the \textit{augmented \method and the original samples}. 

We observe that in low data regimes \figref{fig:dyna_baselines} (right), \algo{CAIAC+MBPO} leads to a  boost in performance compared to \method alone. However, it still underperforms with respect to \code{MBPO}. We hypothesize that this is due to some fraction of unfeasible augmented samples used to train the model, leading to compounding errors when generating the rollout transitions with \code{MBPO}.
This would also support the results for the Fetch-Pick\&Lift environment \figref{fig:dyna_baselines}(left), where despite outperforming \code{MBPO} significantly, \code{CAIAC+MBPO} doesn't achieve same performance as \code{CAIAC} alone.

In terms of implementation details, all the methods share the same hyperparameters, except for the horizon length (N) and the ratio of augmented samples (R) which we tuned individually. We used $N=5$ for \code{CAIAC+MBPO} and $N=10$ for \code{MBPO} and $R=0.5$ for the FetchPush environment, while for FetchPick\&Lift we crucially reduce $R$ to 0.1 for all methods and use $N=5$.

We believe that combining \method with model-based approaches is a very promising direction, and we leave extensive exploration of such an approach for future work.

\end{document}